# Spatial-temporal wind field prediction by Artificial Neural Networks

Jianan Cao[1,*], David J. Farnham[1], Upmanu Lall[1]

[1] Department of Earth and Environmental Engineering, Columbia University, New York, NY

[*] Corresponding author: jc4642@columbia.edu

**Abstract:**

The prediction of near surface wind speed is becoming increasingly vital for the operation of electrical energy grids as the capacity of installed wind power grows. The majority of predictive wind speed modeling has focused on point-based time-series forecasting. Effectively balancing demand and supply in the presence of distributed wind turbine electricity generation, however, requires the prediction of wind fields in space and time. Additionally, predictions of full wind fields are particularly useful for future power planning such as the optimization of electricity power supply systems. In this paper, we propose a composite artificial neural network (ANN) model to predict the 6-hour and 24-hour ahead average wind speed over a large area (~$3.15 \times 10^6$ km$^2$). The ANN model consists of a convolutional input layer, a Long Short-Term Memory (LSTM) hidden layer, and a transposed convolutional layer as the output layer. We compare the ANN model with two non-parametric models, a null persistence model and a mean value model, and find that the ANN model has substantially smaller error than each of these models. Additionally, the ANN model also generally performs better than integrated autoregressive moving average models, which are trained for optimal performance in specific locations.

Key words: Wind speed prediction; Space-time prediction; Artificial Neural Network; Convolution; Long short-term memory (LSTM)

## 1. Introduction

Wind energy has become an increasingly important source of electricity generation, driven by concern over the use of fossil fuels and their associated greenhouse gas emissions and its economic viability in many locations and economies. Increased reliance on wind generators, however, does not come without difficulties. Chief among these is the issue of intermittency and variability, which must be managed by system operators so that load balancing can be constantly achieved. One effective way to manage the variability of wind turbine generator output is through accurate short-term wind speed forecasting.

Wind speed forecasting is crucial for decision making on the scheduling, maintenance, and integration planning. For example, wind forecasts in the range of minutes impact the dispatch operation, forecasts in the range of hours determine the issues of scheduling in a power system, and forecasts in the range of days affect the maintenance and resource planning [24]. Reliable methods and techniques of wind speed forecasting are increasingly important for the characterization and prediction of the wind resource [2] given the rapid development of wind power generation and the increasing integration of wind energy into power systems. In this paper, we address wind speed prediction at the time-scale of hours to days by predicting average wind speed over the next 6 and 24 hours.

The chaotic fluctuation of wind speed makes it difficult to forecast. Achieving high accuracy wind power predictions is thus difficult since wind power is a function of wind speed cubed. Critically, large errors in wind power generation forecasts may cause difficulties in electricity transmission and dispatching. Thus, improving wind speed forecasts is critical for a smooth transition into an era of increased wind power generation. Below we briefly discuss several approaches to wind speed forecasting before presenting our forecasting model.

Mature approaches to wind speed forecasting mainly include physical methods, such as numerical weather forecast (NWF) and mesoscale models [4], conventional statistical methods such as time series models or Markov chain models [5-8], and hybrid physical-statistical models [9]. In recent years, machine learning and deep learning techniques have been adopted for the purpose of wind speed forecasting, such as radial basis functions [12], neural networks (NN) of multi-layer perceptron (MLP) [10,11,22] and recurrent neural networks [13,14,21], fuzzy logic [15,16].

Wind speed prediction using NNs is becoming an increasingly popular topic due in part to the continued advances in computing power. Theoretically speaking, a complex NN can represent any linear or nonlinear function. The inherent nonlinear structure of NNs is useful for managing complex relations in problems of diverse disciplines. NN models can learn from past data, recognize complex patterns or relationships in historical observations and use the relationships to forecast future values [24].

Several NN-based methods to wind speed forecasting have been proposed. Lapedes and Farber [17] proposed a NN model along with a feed forward and error backpropagation algorithm for wind speed forecasting. Song [18] developed a NN-based method to perform one-step-ahead forecast, in which good performance can be obtained when the wind data do not oscillate violently. Alexiadis et al. claimed that the forecasted wind power error of their NN-based prediction model was about 10% lower compared to the persistent error for 10-min and 1-hour forecasting on Kea island [19]. Mehmet Bilgili, Besir Sahin* and Abdulkadir Yasar [20] used a Feedforward Neural Network (FNN) to predict the mean monthly wind speed of any target station using the mean monthly wind speeds of neighboring stations which are indicated as reference stations, the maximum mean absolute percentage error was found to be about 14% for Antakya meteorological station and the best result was found to be 4.5% for Mersin meteorological station [24]. T.G. Barbounis, J.B. Theocharis* [21] suggested a local feedback dynamic fuzzy neural network model to predict multi-step ahead wind speeds from 15 min to 3 h ahead and claimed that the model exhibited superior performance compared to other network types suggested in the literature [21].

Many of NN-based wind speed forecasting models, however, mainly focus on local prediction, i.e. predicting future wind speed at one spatial point based on past values or the past values of its neighborhood points. As noted previously, this point based prediction is a significant shortcoming since proper electrical system management (e.g. load balancing) often requires forecasting of wind speeds over large spatial domains simultaneously. To address this gap, we propose a composite neural network that predicts the 1-step ahead 6-hourly-averaged and 24-hourly-averaged distributed wind speed over a large domain, which covers about half of the continental United States of America simultaneously.

## 2. Data and Methods

### 2.1 Data

The hourly average wind speed at 10 m above displacement height is the primary dataset used in this study. The data was collected from NASA GSFC MERRA through the IRI Data Library. The target area we selected is a rectangular domain in the United States (U.S.) whose latitude expands from 30.0N to 45.5N, and longitude expands from 84.67W to 105.33W. The grid resolution is about 0.67° (longitude) by

0.5° (latitude), consequently we get a 32×32 matrix in space. The time length of the data set is 324336 hours (37 years), from Jan 1$^{st}$, 1979 to Dec 31$^{st}$, 2015.

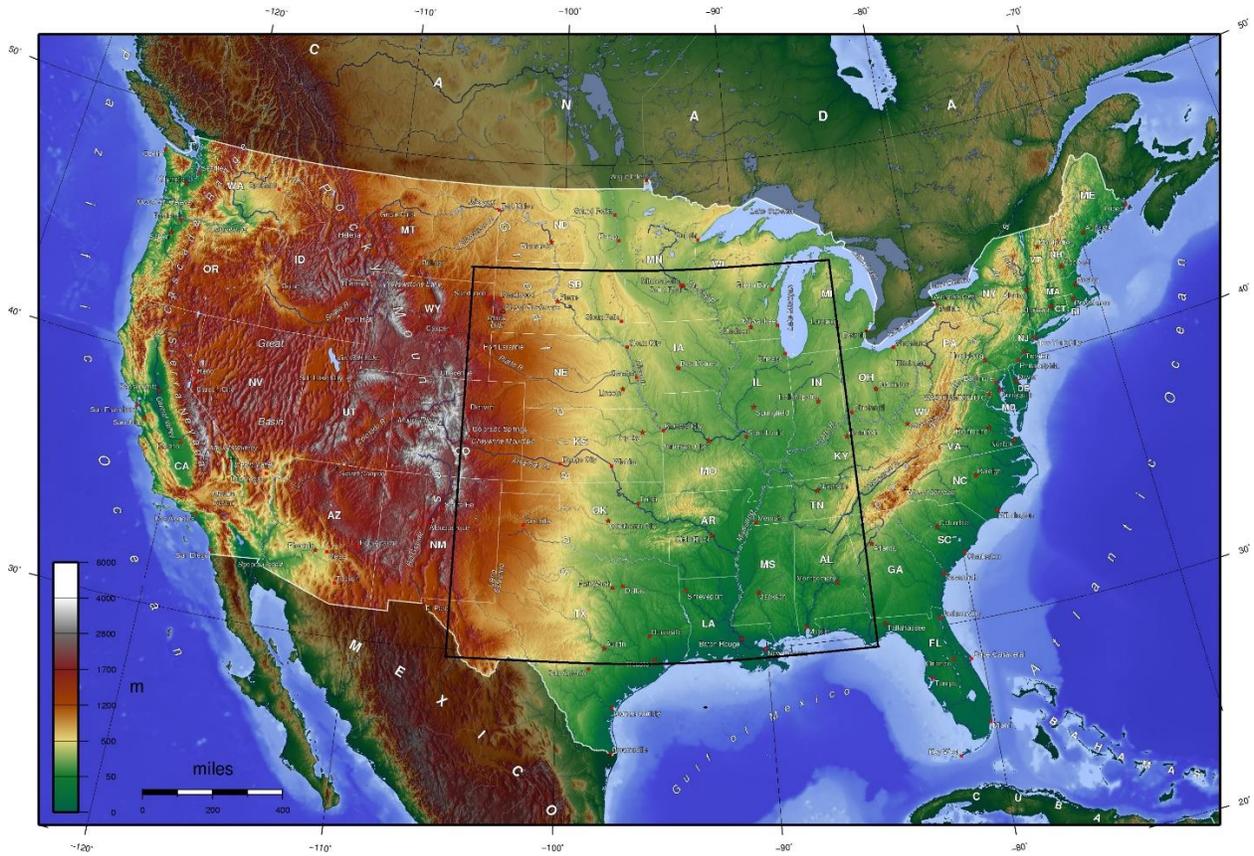

**Fig. 2.1.1 The topography of our target area. The black box shows the domain over which the analysis and wind speed prediction is conducted.**

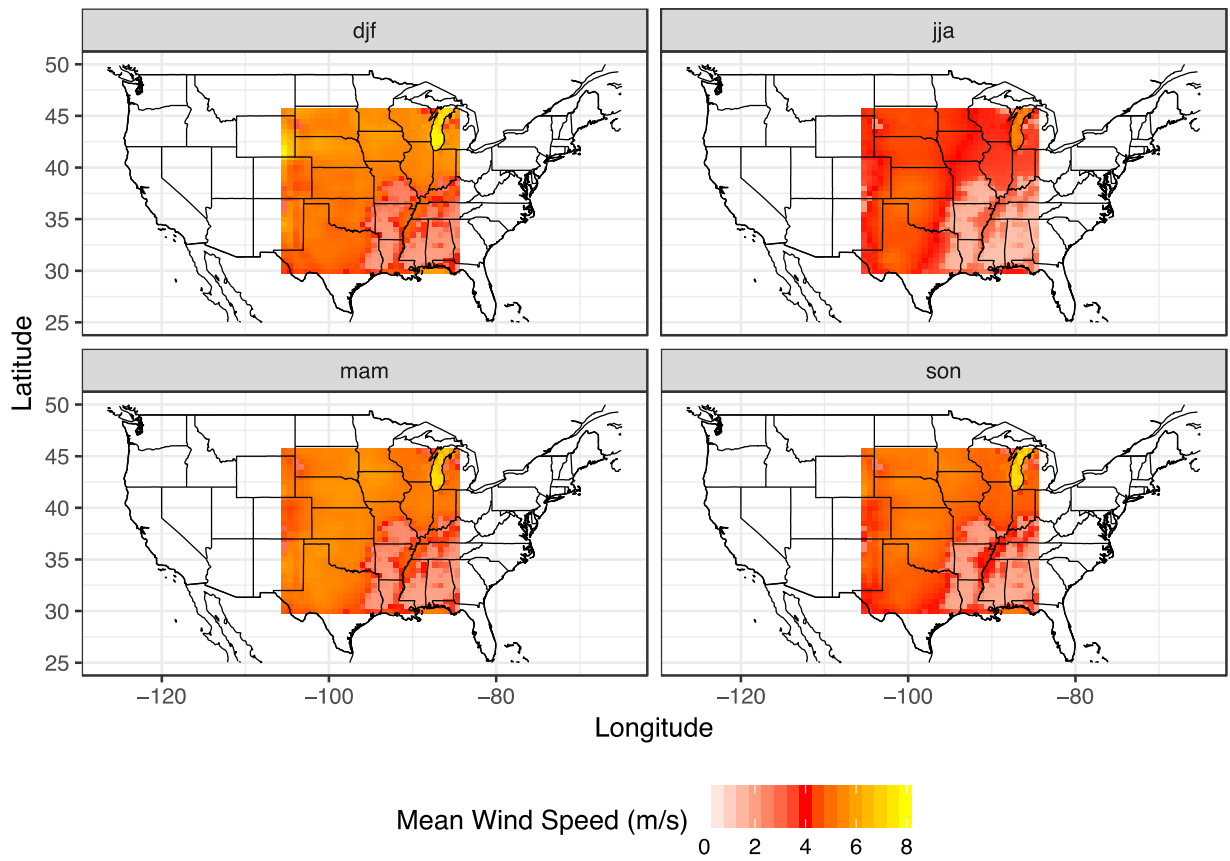

Fig. 2.1.2 The mean of wind speed data in our target area

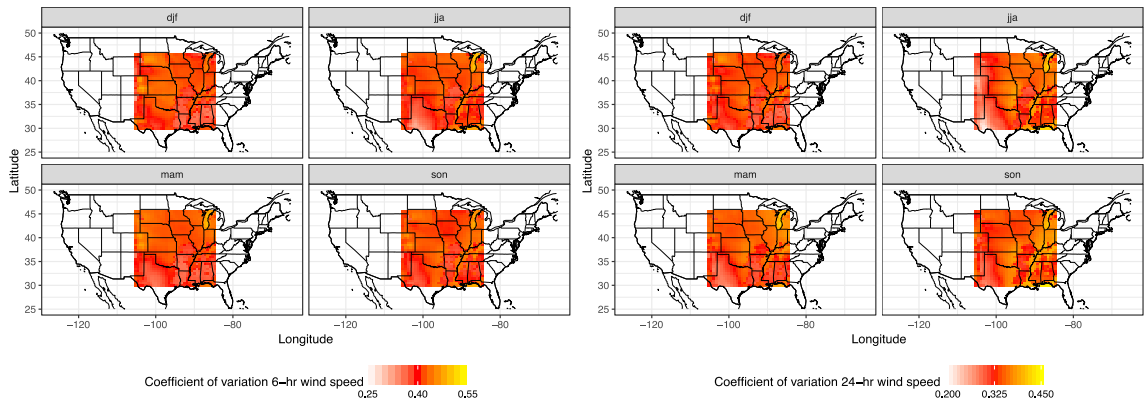

Fig. 2.1.3 coefficient of variation (standard deviation divided by mean) for 6-hour-averaged (left) and 24-hour-averaged (right) wind speed in our target area

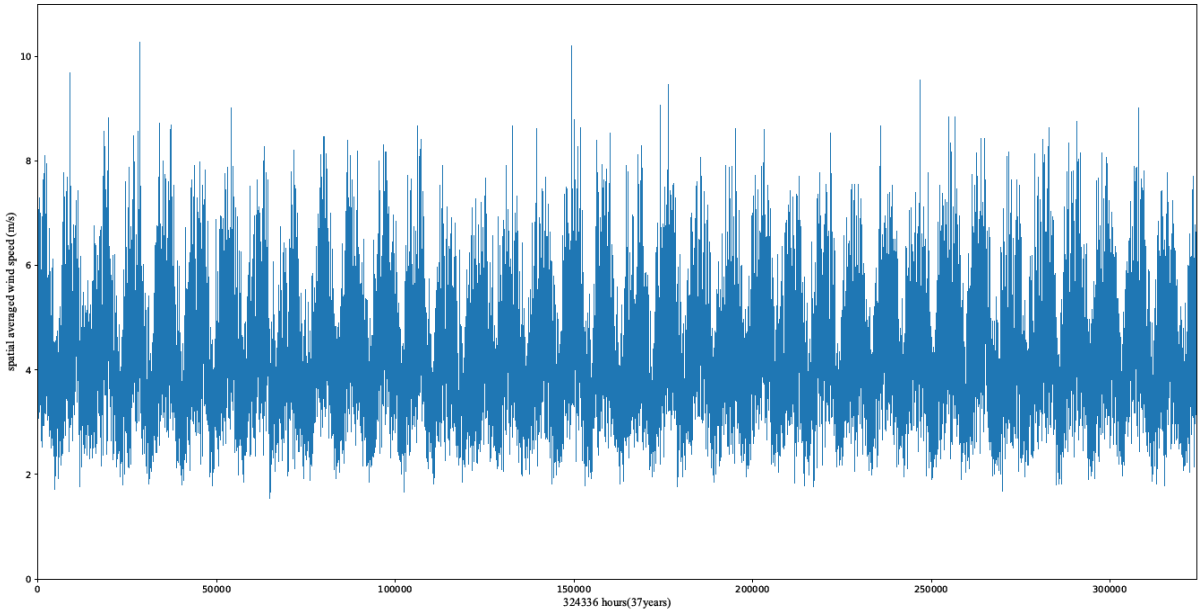

**Fig. 2.1.4 Time series of spatially averaged hourly wind speed**

Figures 2.1.1-2.1.3 illustrate that the mean and coefficient of variation of wind speed are highly correlated with topography and season. The highest wind speeds and coefficients of variation in the study region are over Lake Michigan, while low wind speed are present over low altitude plans in the southeastern portion of the study region. The boreal summer JJA (Jun-Aug) season has the lowest wind speeds in the study area, while the DJF (Dec-Feb) and MAM (Mar-May) seasons have the highest wind speeds. Clear annual periodicity of the spatially averaged wind speed is evident in Fig. 2.1.4. Lastly, wind speeds at nearby points also perform strong correlations with each other. We ordered the 1024 spatial points by row (the most northwest point as the $0^{th}$, and the most southeast point as the $1023^{rd}$), then computed the correlations between these points (Fig. 2.1.5).

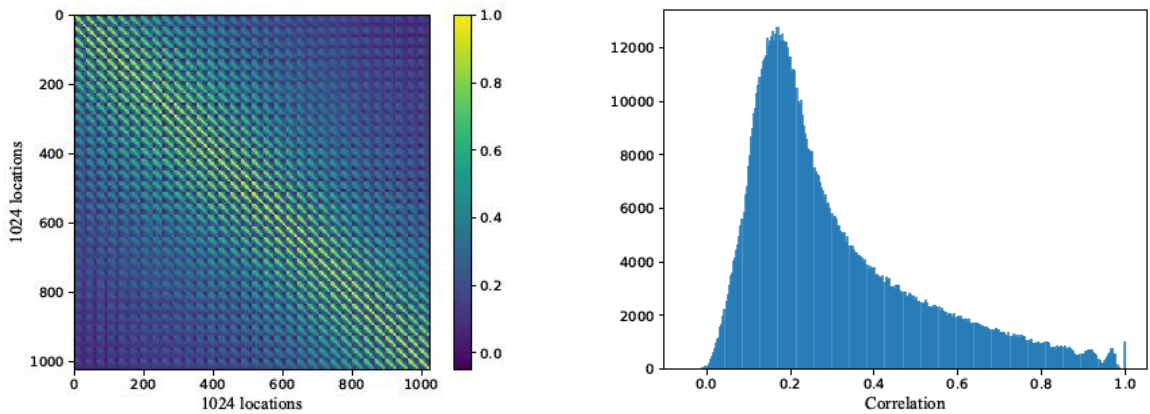

**Fig. 2.1.5 Spatial correlation matrix and distribution histogram for hourly wind speed**

## 2.2 Introduction to the structure of our ANN

The Neural Network employed in this paper consists of three layers:

(1) one convolutional layer with kernel size (3,3); this is the input layer,
(2) one LSTM recurrent layer; i.e. the hidden layer,
(3) one transposed convolutional layer; the output layer.

The neural networks are constructed in Python 3.5 environment using Keras [27] (The Python Deep Learning Library) with Tensorflow [26] as the backend.

### 2.2.1 Convolutional Layer

Convolutional layers apply a convolution operation to the input data and pass the result to the next layer. 2-D convolution is widely used in image processing. Given a picture $x_{ij}$ ($1 \leq i \leq M, 1 \leq j \leq N$) and a filter $f_{ij}$ ($1 \leq i \leq m, 1 \leq j \leq n$), typically $m \ll M, n \ll N$, the output of convolution is:

$$y_{ij} = \sum_{u=1}^{m} \sum_{v=1}^{n} f_{uv} \cdot x_{i-u+1,j-v+1}$$

Just as the value of each pixel in an image is always related to the values of its surrounding pixels, the wind speed at each spatial point is also correlated with its neighborhood points (Fig. 2.1.5). Consequently, inspired by image processing techniques, here we also introduce a convolution layer to capture the features of wind speed behavior in our two-dimensional region. The activation function used in the convolutional layer is a rectified linear unit (ReLU) function:

$$f(x) = \max(0, x)$$

which transforms the activation values of the neurons in convolutional layer to the input of next layer.

### 2.2.2 Long Short-Term Memory (LSTM) recurrent layer

In many real tasks, the input has temporal dependence, such as video, voice, text, and other sequential data structures. And the output at a certain moment may be related to the input at the previous moment. So does our spatial-temporal wind field, the wind speed at one spatial point might be related to the wind speed at its neighborhood points at the previous moment. The recurrent neural networks (RNN) can handle sequences of arbitrary length by using neurons with self-feedback. The RNN is more in line with the structure of the bioneural network.

Given an input sequence $x_{1:T} = (x_1, x_2, \dots, x_t, \dots, x_T)$, a recurrent neural network (RNN) can update the active value $h_t$ of the hidden layer with feedback edges by the following formula:

$$h_t = \begin{cases} 0 & t = 0 \\ f(h_{t-1}, x_t) & otherwise \end{cases} \quad (2.2.1)$$

Mathematically speaking, the formula (2.2.1) can be regarded as a dynamic system. Consequently, $h_t$ is called state or hidden state in many articles. Theoretically, RNN can be used to simulate any dynamic system.

Long Short-Term Memory neural network (LSTM) is one of the most efficient version of RNN. When the input sequence is too long, it could efficiently avoid gradient exploding or vanishing problems when we use the Backpropagation Through Time (BPTT) algorithm to learn the parameters of RNNs.

The key point of LSTM model is using internal Memory Units to storage history information. Then let the network learn when to forget the historical information and when to update the memory units using new information dynamically. At time t, the internal memory unit $c_t$ records all historical information up to the current moment and is controlled by three gates: the input gate $i_t$, the 'forget' gate $f_t$, and the output gate $o_t$. And the values of the elements of the three gates vary in [0,1]. 0 represents close state, no information could pass; and 1 represents open state, all the information can pass. 'forget' gate controls how much information each unit needs to forget, input gate controls how much information is added to each memory cell, and the output gate controls how much information is output per memory unit.

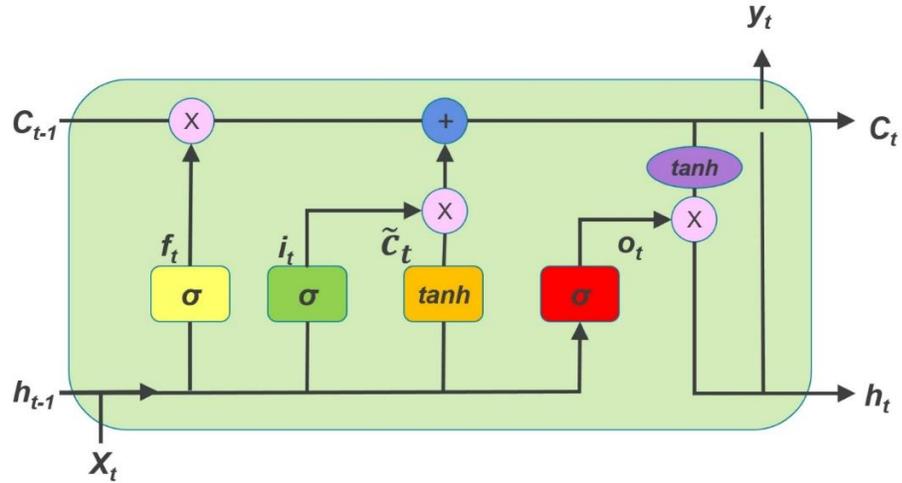

Fig. 2.2.1 The inner structure of LSTM [1]

The update equations at time t are:

$$i_t = \sigma(W_i x_t + U_i h_{t-1} + b_i),$$

$$f_t = \sigma(W_f x_t + U_f h_{t-1} + b_f),$$

$$o_t = \sigma(W_o x_t + U_o h_{t-1} + b_o),$$

$$\tilde{c}_t = tanh(W_c x_t + U_c h_{t-1} + b_c),$$

$$c_t = f_t \odot c_{t-1} + i_t \odot \tilde{c}_t,$$

$$h_t = o_t \odot tanh(c_t)$$

Where $x_t$ is the input vector at time t, σ is the logistic function, W, U are coefficient matrixes, b is the bias vector, ⊙ is the Hadamard product.

Due to this special structure, the network can potentially learn long-term system behavior. LSTM is widely used in many tasks, such as machine translation. The connection of cells is shown as below:

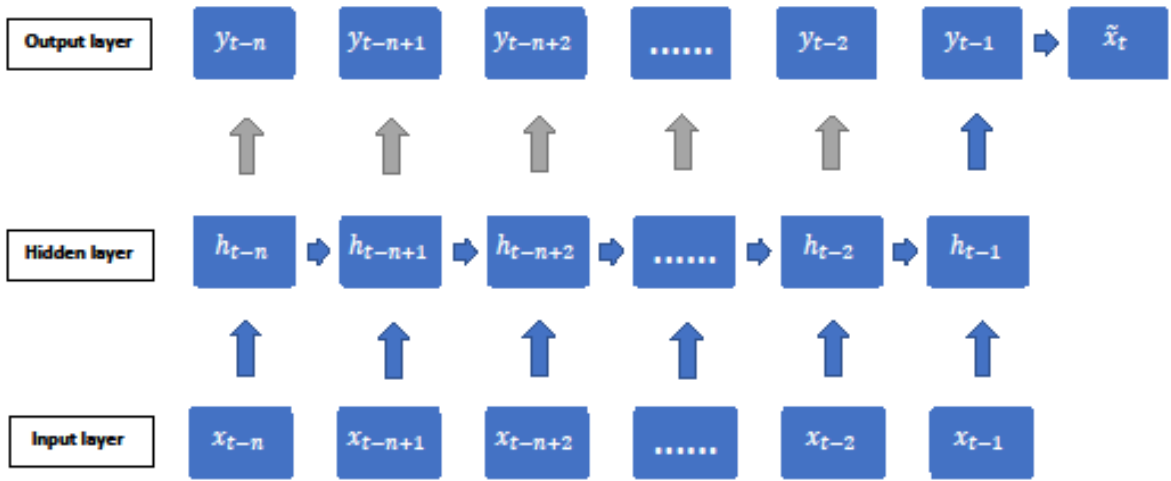

**Fig. 2.2.2 macrostructure of recurrent layer**

Since we are using the wind speed at time $t-n, t-n+1, \ldots, t-1$ to forecast the wind speed at time t, we do not care the performance of outputs $y_{t-2}, y_{t-3}, \ldots, y_{t-n+1}, y_{t-n}$ and only pay attention to the performance of $y_{t-1}$, which is equal to $\tilde{x}_t$, the prediction for $x_t$. Finally, we pass the result $\tilde{x}_t$ to the output layer (i.e. the transposed convolutional layer).

### 2.2.3 Transposed convolution layer

The need for transposed convolutions generally arises from the desire to use a transformation going in the opposite direction of a normal convolution, i.e., from something that has the shape of the output of some convolution to something that has the shape of its input while maintaining a connectivity pattern that is compatible with said convolution [25]. For instance, one might use such a transformation as the decoding layer of a convolutional auto-encoder or to project feature maps to a higher-dimensional space [25]. Here we need to transform the 30x30 value matrix which comes from convolutional layer to a 32x32 value matrix, which represents the wind speed field at time t.

One efficient way to realize the transposed convolution works by swapping the forward and backward passes of a convolution, and then compute the convolution kernel.

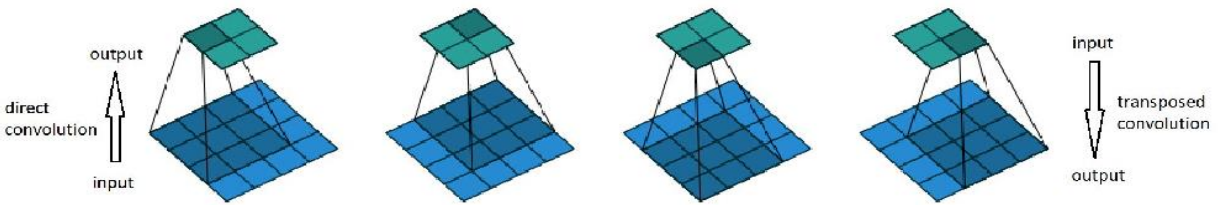

**Fig. 2.2.3 The transpose of convolving a 3×3 kernel over a 4×4 input using unit strides. It is equivalent to convolving a 3×3 kernel over a 2×2 input padded with a 2×2 border of zeros using unit strides [25].**

Let's consider the transpose of convolving a 3×3 kernel over a 4×4 input using unit strides. If the input and output were to be unrolled into vectors from left to right, top to bottom, the convolution could be represented as a sparse matrix $C$ where the non-zero elements are the elements $\omega_{i,j}$ of the kernel (with $i$ and $j$ being the row and column of the kernel respectively) [25]:

$$\begin{pmatrix} \omega_{0,0} & \omega_{0,1} & \omega_{0,2} & 0 & \omega_{1,0} & \omega_{1,1} & \omega_{1,2} & 0 & \omega_{2,0} & \omega_{2,1} & \omega_{2,2} & 0 & 0 & 0 & 0 & 0 \\ 0 & \omega_{0,0} & \omega_{0,1} & \omega_{0,2} & 0 & \omega_{1,0} & \omega_{1,1} & \omega_{1,2} & 0 & \omega_{2,0} & \omega_{2,1} & \omega_{2,2} & 0 & 0 & 0 & 0 \\ 0 & 0 & 0 & 0 & \omega_{0,0} & \omega_{0,1} & \omega_{0,2} & 0 & \omega_{1,0} & \omega_{1,1} & \omega_{1,2} & 0 & \omega_{2,0} & \omega_{2,1} & \omega_{2,2} & 0 \\ 0 & 0 & 0 & 0 & 0 & \omega_{0,0} & \omega_{0,1} & \omega_{0,2} & 0 & \omega_{1,0} & \omega_{1,1} & \omega_{1,2} & 0 & \omega_{2,0} & \omega_{2,1} & \omega_{2,2} \end{pmatrix}$$

This linear operation takes the input matrix flattened as a 16-dimensional vector and produces a 4-dimensional vector that is later reshaped as the 2×2 output matrix. Using this representation, we have:

$$C \cdot X^{(16)} = Y^{(4)} \tag{2.2.2}$$

$$X^{(16)} = (C^T C)^{-1} C^T Y^{(4)} \tag{2.2.3}$$

Where $X^{(16)}$ is the 16-dimensional input vector and $Y^{(4)}$ is the 4-dimensional output vector. Since matrix $C^T C$ is always singular, typically we cannot get $(C^T C)^{-1}$. Consequently, we need to revise the Eq. (2.2.3) into a regularized form:

$$X^{(16)} = (C^T C + \lambda I)^{-1} C^T Y^{(4)} \tag{2.2.4}$$

Where $\lambda$ is the regularization parameter, $I$ is the identity matrix. Actually Eq. (2.2.4) is the optimized solution of the regularization problem:

$$\boldsymbol{min.} \quad Loss(X^{(16)}) = \|C X^{(16)} - Y^{(4)}\|_2^2 + \lambda \|X^{(16)}\|_2^2$$

$$\boldsymbol{s.t.} \quad X_i^{(16)} \geq 0, \quad i = 1, 2, \dots, 16$$

The activation function used in the transposed convolutional layer is a linear function:

$$f(x) = x$$

The methods for transposed convolution used in our research is the default method in the tensorflow class ***tf.keras.layers.Conv2DTranspose***.

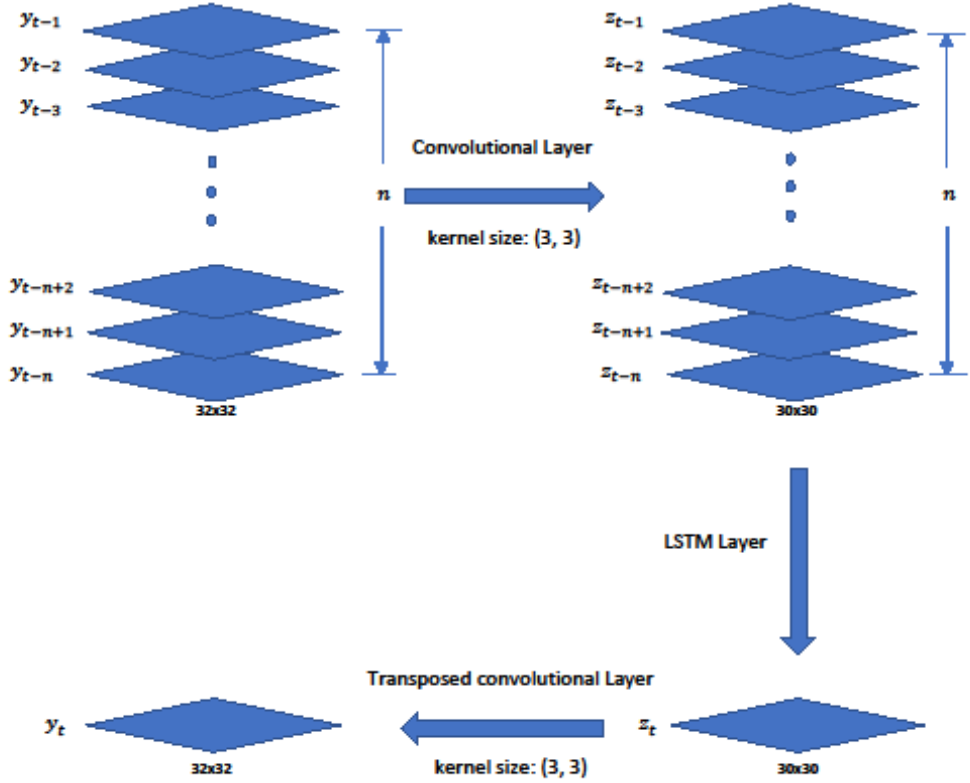

**Fig. 2.2.5 Sketch of ANN**

## 3. Simulation

### 3.1 Constructing wind speed data sequences:

We used the averaged wind speed of past n continuous 6-hour(24-hour) blocks to predict the averaged wind speed of the next 6-hour(24-hour) block. So we first need to construct the sequences of blocks.

#### 6-hour-averaged wind speed:

Here our 37-year-long dataset is divided into three parts: the first 10 percent as the test set, the last 10 percent as the validation set, the rest 80 percent as the training set. The first thing we need to do is to construct the 6-hour-averaged wind speed data sequence $X_{t-6i}(i = 1,2,...,n)$ and corresponding label $X_t$, where $X_k(t = 1,2,...)$ is the averaged wind speed on the 6-hour period from time $k$ to time $k + 5$. we randomly select time points from our data sets, then around each time point $t$ $(t \geq 6n + 1, t \leq T - 5, \ T \ is \ the \ length \ of \ raw \ data \ set)$, we compute the data sequences and label by:

$$X_{t-6i} = \frac{1}{6}\sum_{k=0}^{5} Z_{t-6i+k} \quad i = 0,1,2,...,n$$

Where $Z_t, t = 1,2,...,T$ is the hourly averaged raw wind speed. The sample rate of time points is close to 1.0 in both validation set and test set, and 0.8 in training set. The $n$ for 6-hour-averaged speed simulation is 6, i.e. using the data of past 36 hours to predict the mean wind speed of next 6 hours.

**24-hour-averaged wind speed :**

Here our 37-year-long dataset is divided into three parts: the first 5 percent as the validation set, the last 5 percent as the test set, the rest 90 percent as the training set. Similar with 6-hour mean, we calculate the data sequences and labels for 24-hour-averaged wind speed by:

$$X_{t-24i} = \frac{1}{24} \sum_{k=0}^{23} Z_{t-24i+k} \qquad i = 0,1,2,\dots,n$$

Where $Z_t, t = 1,2,\dots,T$ is the hourly averaged raw wind speed. The sample rate of time points is 1 in validation set and test set, and 0.9 in training set. The $n$ for 24-hour-averaged speed simulation is 3, i.e. using the data of past 72 hours to predict the mean wind speed of next 24 hours.

## 3.2 Preprocessing:

### Linear shift:

Before importing data into our ANN, we use a linear shift to approximately transfer the wind speed data into the range [0,1]:

$$Y_t = X_t/\beta \tag{3.2.1}$$

Where $\beta$ is the maximal wind speed in our training set, $X_t$ is transformed sequences of 32x32 wind speed data matrix and $Y_t$ is the input of ANN.

Corresponding to the linear shift at input, we apply an inversed linear shift on the output of our ANN:

$$\hat{X}_t = \beta \cdot \tilde{Y}_t \tag{3.2.2}$$

Here $\beta$ is the constant we divide in preprocessing, $\tilde{Y}_t$ is the output of ANN and $\hat{X}_t$ is the prediction for 6-hour or 24-hour averaged wind speed $X_t$.

Here we do not add any other preprocessing method (like extracting the mean, divided by standard deviance, etc.) except a linear shift on the input data, because we do not want to put much priori assumption on our model. Unlike other classical statistical models, Neural Network can almost simulate any kinds of function. We hope the ANN models could find the implied relationship among the wind speed only by the huge original data set and the complex network structure.

## 3.3 Optimization with regularization:

Here we assume the target wind speed is

$$Y = f(x) + \varepsilon \tag{3.3.1}$$

$f(x)$ is a complex and unknown function, $\varepsilon$ is a random noise with zero mean. Now we want to find a function $\hat{Y} = \hat{f}(x)$ to simulate the behavior of $f(x)$. However, we don't know what $f(x)$ looks like, and the non-linear function $\hat{f}(x)$ built by ANN has amounts of coefficients, which can simulate a complex enough function. If we directly set the optimization problem to be minimizing $\|\hat{Y} - Y\|$, then our model $\hat{f}(x)$ will be trained to emulate the behavior of $f(x) + \varepsilon$, which would cause overfitting. To avoid this

problem, we need to add a regularization term to the loss function, which sacrifices some accuracy but helps get a more stable solution. Consequently, the estimation we got from ANN is neither an unbiased estimation, nor an MLE estimation because of the shrinkage.

Here we choose the L2 distance $\left\|\hat{Y} - Y\right\|_2^2$ to measure the deviation of our prediction, i.e., select the mean squared error as the loss function of ANN.

$$6 - hour: \quad Loss = \frac{1}{T-6n-5}\sum_{t=6n+1}^{T-5}\left\|Y_t - \hat{Y}_t\right\|_2^2 + \lambda \sum_i \|\beta_i\|_2^2 \tag{3.3.2}$$

$$24 - hour: \quad Loss = \frac{1}{T-24n-23}\sum_{t=24n+1}^{T-23}\left\|Y_t - \hat{Y}_t\right\|_2^2 + \lambda \sum_i \|\beta_i\|_2^2 \tag{3.3.3}$$

The first term of RHS in both equations (3.3.2) and (3.3.3) is the mean squared error on training set, the second term is the L2 penalty on model coefficients. The regularization constant $\lambda$ is set to be 0.15 in both two loss functions.

Both in 6-hour and 24-hour wind speed modeling, the optimizer used in ANN fitting is RMSprop. RMSprop is an optimizer that utilizes the magnitude of recent gradients to normalize the gradients. We always keep a moving average over the root mean squared (hence RMS) gradients, by which we divide the current gradient. Let $f'(\theta_t)$ be the derivative of the loss with respect to the parameters at time step $t$. In its basic form, given a learning rate $\alpha$ (0.001 as default) and a decay term $\gamma$ (0.9 as default) we perform the following updates [3]:

$$r_t = (1-\gamma)f'(\theta_t)^2 + \gamma r_{t-1},$$

$$v_{t+1} = \frac{\alpha}{\sqrt{r_t}}f'(\theta_t),$$

$$\theta_{t+1} = \theta_t - v_{t+1}.$$

In some cases, adding a momentum term $\beta$ is beneficial. Here, Nesterov momentum is used:

$$\theta_{t+\frac{1}{2}} = \theta_t - \beta v_t,$$

$$r_t = (1-\gamma)f'\left(\theta_{t+\frac{1}{2}}\right)^2 + \gamma r_{t-1},$$

$$v_{t+1} = \beta v_t + \frac{\alpha}{\sqrt{r_t}}f'\left(\theta_{t+\frac{1}{2}}\right),$$

$$\theta_{t+1} = \theta_t - v_{t+1}$$

RMSprop is a very robust optimizer which has pseudo curvature information. Additionally, it can deal with stochastic objectives very nicely, making it applicable to mini batch learning [3]. All the coefficients in RMSprop optimizer are set as default in our research.

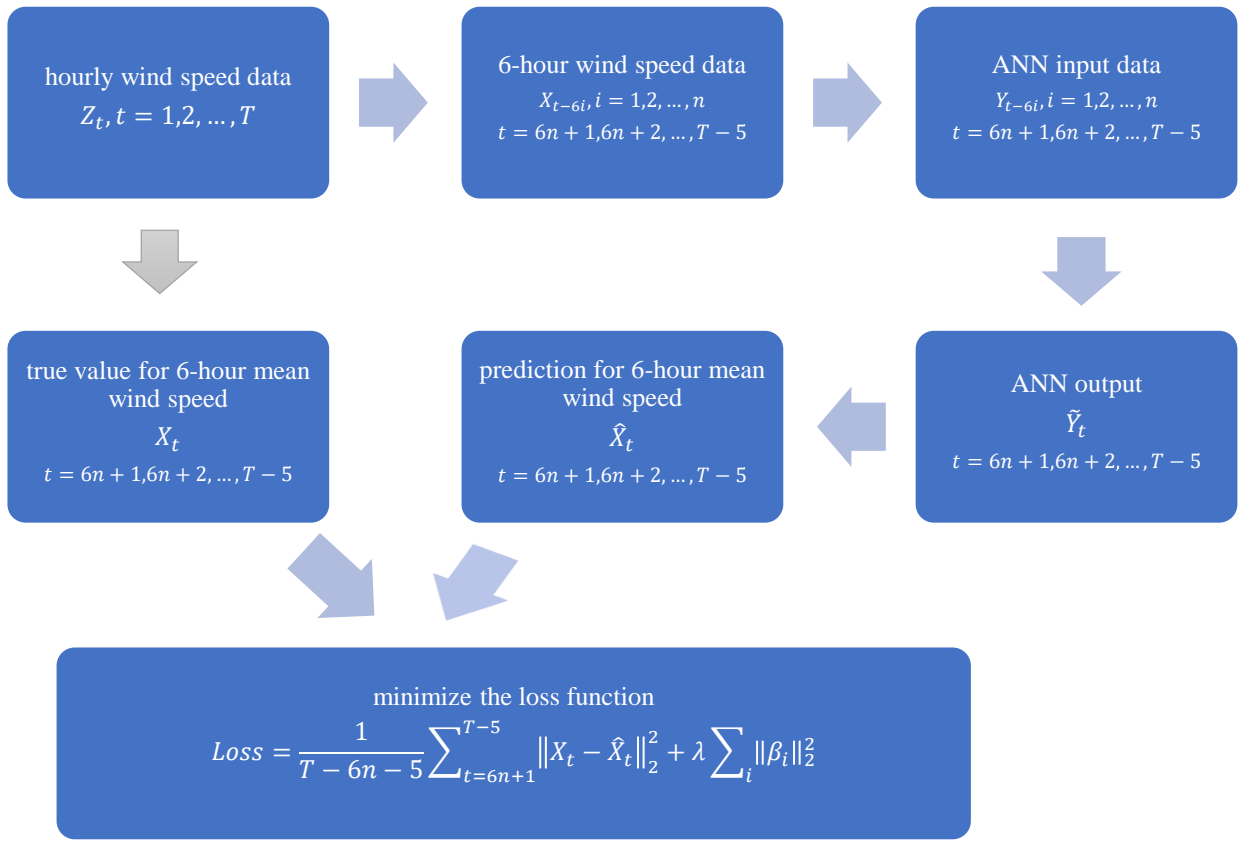

Fig. 3.1 sketch of 6-hour ANN fitting process

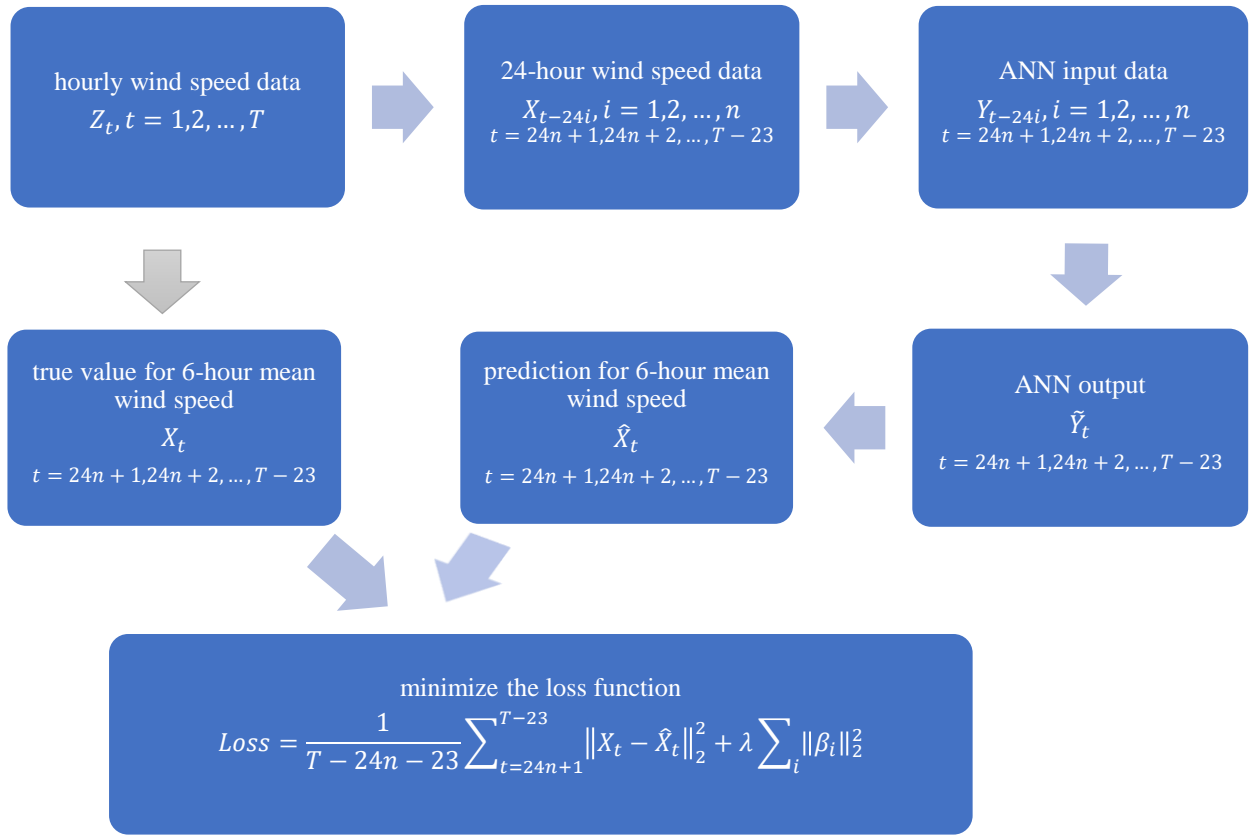

Fig. 3.2 sketch of 24-hour ANN fitting process

## 4. Results and Error Analysis:

Our results indicate that in almost all the locations and during all seasons, our ANN model outperforms the null persistence model and the mean value model. We also pick specific locations to test the performance of the ANN model against that of an optimally chosen ARIMA model. Again, the ANN model has better performance.

### 4.1 ANN versus Persistence Model

The ANN model is pointwise compared with persistence model on test set. Based on the test error, two kinds of MSE (MAE) array are computed: MSE (MAE) sequence and MSE (MAE) matrix. The MSE (MAE) sequence is a time series of spatial mean squared error from 1024 different locations. The MSE (MAE) matrix is a 32×32 error matrix in which each entry is the mean squared error in time at that location. Results for 6-hour-averaged and 24-hour-averaged wind speed prediction are shown in Fig. 4.1.1 and Fig. 4.1.2, a statistical summary is exhibited in Table 1.

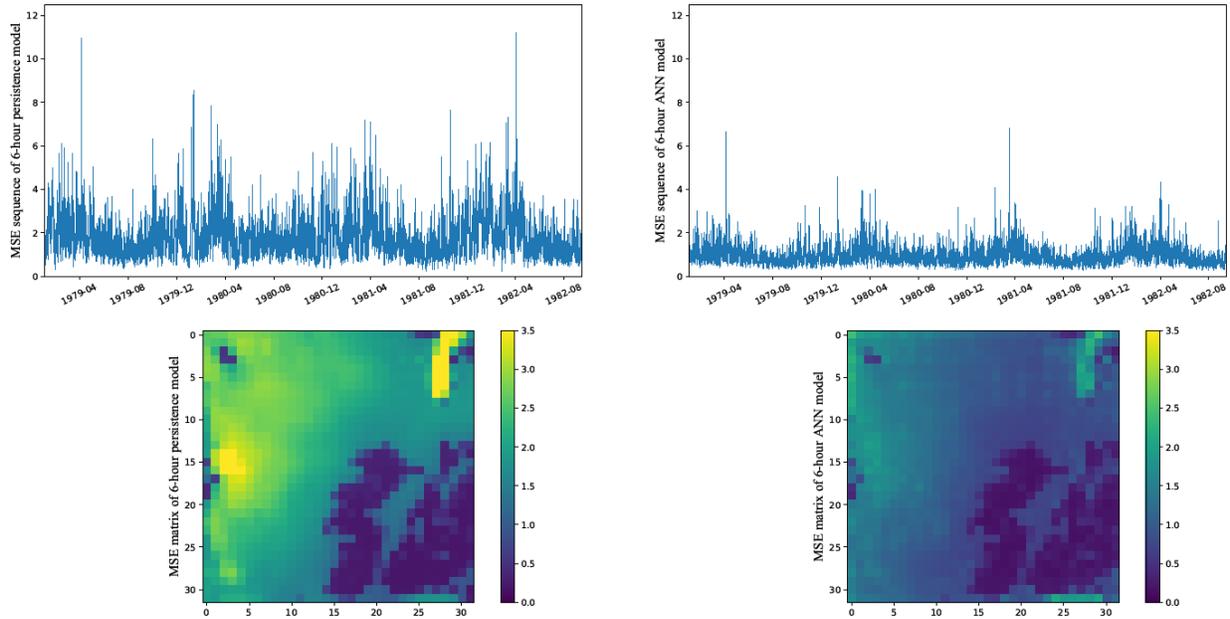

**Fig. 4.1.1 MSE comparison between persistence model (left) and ANN (right) in 6-hour-averaged wind speed prediction**

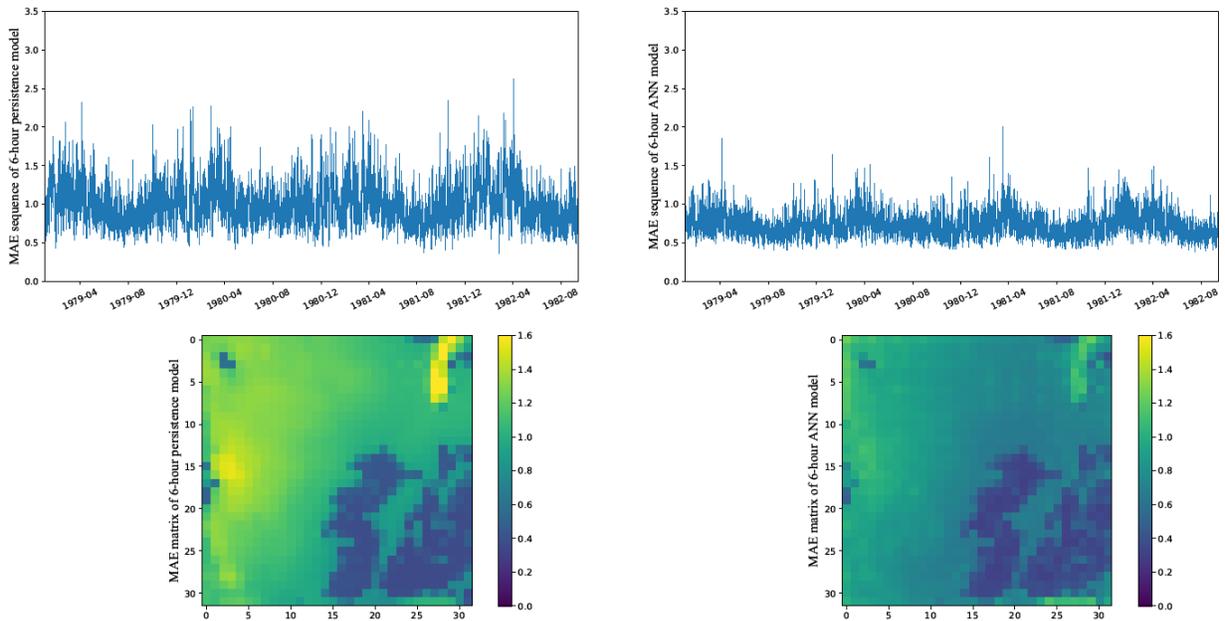

**Fig. 4.1.2 MAE comparison between persistence model (left) and ANN (right) in 6-hour-averaged wind speed prediction**

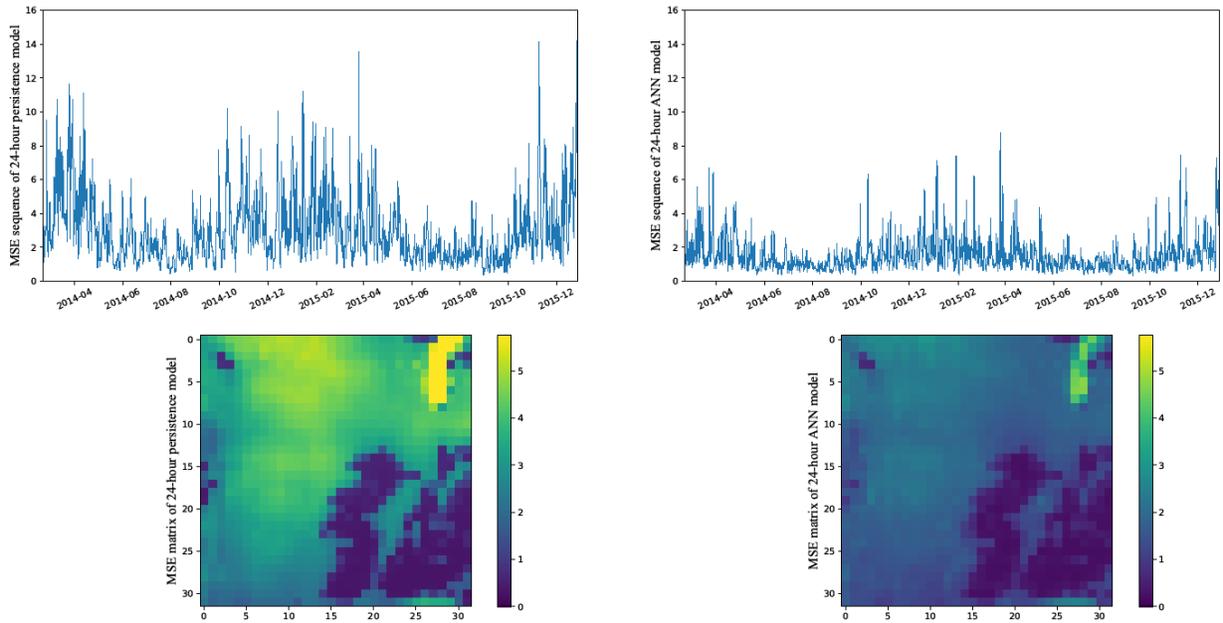

**Fig. 4.1.3 MSE comparison between persistence model (left) and ANN (right) in 24-hour-averaged wind speed prediction**

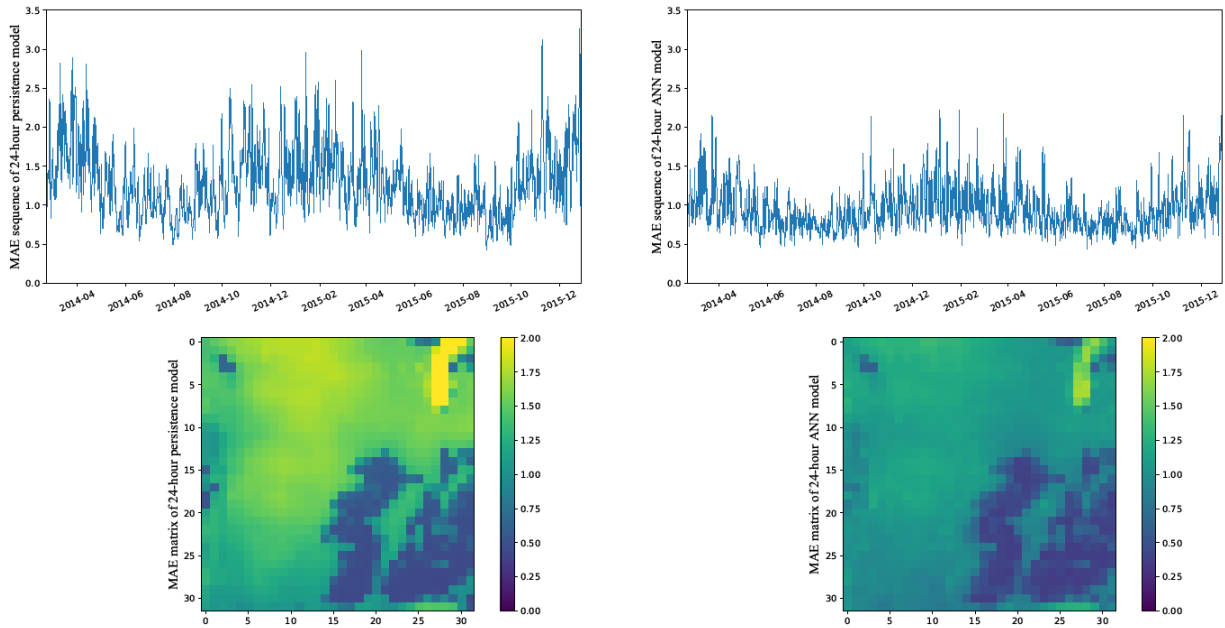

**Fig. 4.1.4 MAE comparison between persistence model (left) and ANN (right) in 24-hour-averaged wind speed prediction**

**Table 1. statistical summary of MSE comparison between ANN and persistence model for 6-hour-averaged and 24-hour-averaged wind speed prediction (unit: $m^2/s^2$)**

|  | MSE in 6-hour-wind models | | | | MSE in 24-hour-wind models | | | |
|---|---|---|---|---|---|---|---|---|
|  | persistence | | ANN | | persistence | | ANN | |
|  | Sequence | Matrix | Sequence | Matrix | Sequence | Matrix | Sequence | Matrix |
| Max | 11.2073 | 4.3994 | 6.8226 | 2.4461 | 14.1695 | 10.6154 | 8.7682 | 4.7204 |
| Min | 0.2304 | 0.2055 | 0.2827 | 0.1508 | 0.3833 | 0.3097 | 0.3091 | 0.2025 |
| Median | 1.5340 | 1.8776 | 0.8496 | 0.9023 | 2.4281 | 3.2691 | 1.2720 | 1.6860 |
| Mean | 1.7726 | 1.7726 | 0.9384 | 0.9384 | 2.9707 | 2.9707 | 1.5482 | 1.5482 |
| Standard Deviance | 0.9544 | 0.9287 | 0.4067 | 0.4771 | 1.9511 | 1.6327 | 0.9557 | 0.7657 |

**Table 2. statistical summary of MAE comparison between ANN and persistence model for 6-hour-averaged and 24-hour-averaged wind speed prediction (unit: $m/s$)**

|  | MAE in 6-hour-wind models | | | | MAE in 24-hour-wind models | | | |
|---|---|---|---|---|---|---|---|---|
|  | persistence | | ANN | | persistence | | ANN | |
|  | Sequence | Matrix | Sequence | Matrix | Sequence | Matrix | Sequence | Matrix |
| Max | 2.6209 | 1.6233 | 2.0001 | 1.2509 | 3.2585 | 2.5702 | 2.2200 | 1.7523 |
| Min | 0.3572 | 0.3564 | 0.3899 | 0.3039 | 0.4310 | 0.4322 | 0.4371 | 0.3588 |
| Median | 0.9460 | 1.0682 | 0.7068 | 0.7387 | 1.2151 | 1.4096 | 0.8863 | 1.0289 |
| Mean | 0.9911 | 0.9911 | 0.7279 | 0.7279 | 1.2837 | 1.2837 | 0.9457 | 0.9457 |
| Standard Deviance | 0.2664 | 0.3242 | 0.1480 | 0.2111 | 0.4328 | 0.4326 | 0.2679 | 0.2752 |

In 6-hour-averaged wind speed prediction, the RMSE of ANN model ($\sqrt{0.9384\,m^2/s^2} \approx 0.9687\,m/s$) is 27.24% less than that of the persistence model ($\sqrt{1.7726\,m^2/s^2} \approx 1.3314\,m/s$), the global MAE of ANN model (0.7279 m/s) is 26.56% less than that of the persistence model (0.9911 m/s).

In 24-hour-averaged wind speed prediction, RMSE of ANN model ($\sqrt{1.5482\,m^2/s^2} \approx 1.2443\,m/s$) is 27.81% less than that of the persistence model ($\sqrt{2.9707\,m^2/s^2} \approx 1.7236\,m/s$), the global MAE of ANN model (0.9457 m/s) is 26.33% less than that of the persistence model (1.2837 m/s).

The relative error, absolute error divided by wind speed magnitude, is also computed. From the perspective of relative error, ANN is also better. The relative error matrix and sequence are shown as below.

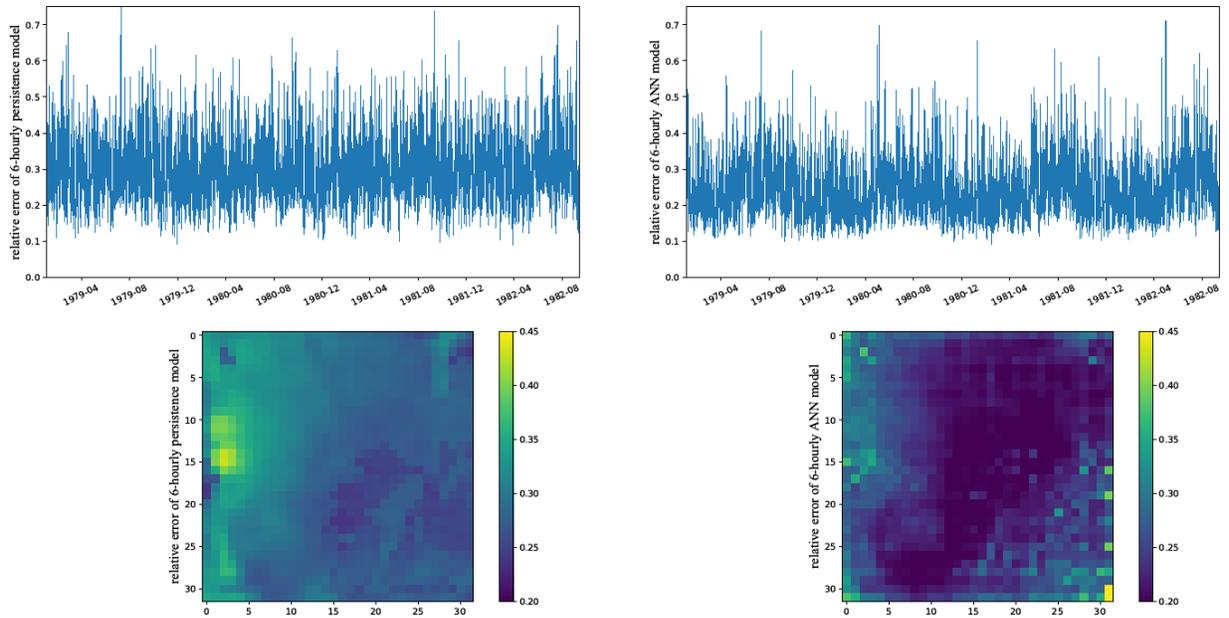

**Fig. 4.1.5 Comparison of mean relative error between persistence model (left) and ANN (right) in 6-hour-averaged wind speed prediction**

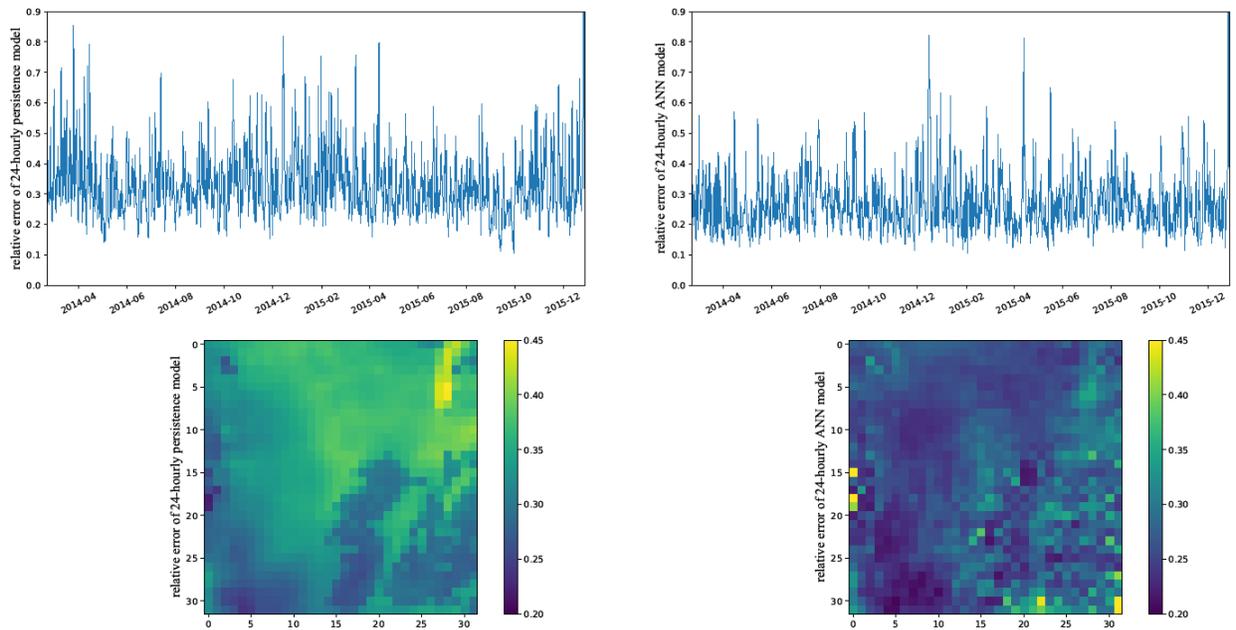

**Fig. 4.1.6 Comparison of mean relative error between persistence model (left) and ANN (right) in 24-hour-averaged wind speed prediction**

**Table 3. Comparison of relative error between persistence model and ANN at different quantiles**

|  | Relative Error | | | |
|---|---|---|---|---|
|  | 6-hour | | 24-hour | |
| Quantile | persistence | ANN | persistence | ANN |
| 0.025 | 0.0086 | 0.0063 | 0.0116 | 0.0087 |
| 0.25 | 0.0885 | 0.0654 | 0.1187 | 0.0889 |
| 0.5 | 0.1966 | 0.1435 | 0.2535 | 0.1887 |
| 0.75 | 0.3732 | 0.2706 | 0.4357 | 0.3363 |
| 0.975 | 1.2035 | 1.1508 | 1.2320 | 1.0766 |

### 4.2 ANN versus mean value model

From Fig. 2.1.4 we know that the wind speeds change significantly during different month. Here we generated another control model------mean value model, which uses the mean wind speed of each month (January, February, …, December) on the training set as the prediction------to compare with the ANN model. Similar with section 4.1, here we also compute the MSE(MAE) sequence and MSE(MAE) matrix.

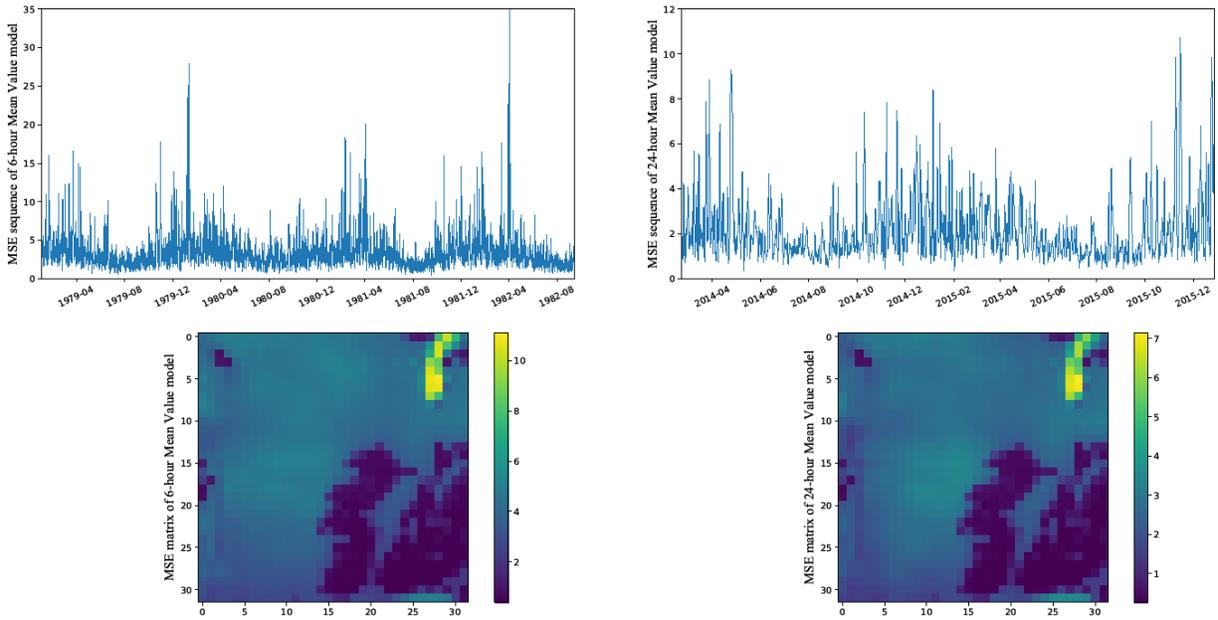

**Fig. 4.2.1 MSE for mean value model in 6-hour-averaged (left) and 24-hour-averaged (right) wind speed prediction**

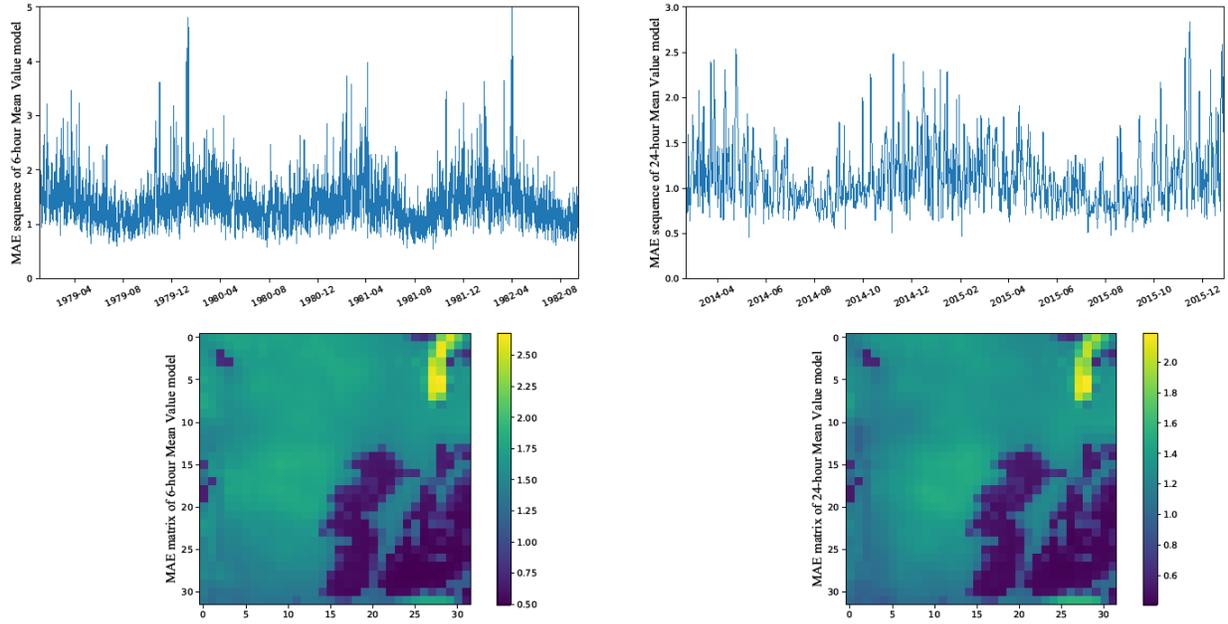

**Fig. 4.2.2 MAE for mean value model in 6-hour-averaged (left) and 24-hour-averaged (right) wind speed prediction**

Compared the two Figures above with Figures 4.1.1 to 4.1.4, we find that the persistence model outperforms the mean value model in 6-hour-averaged wind speed prediction, while that the mean value model outperforms the persistence model in 24-hour-averaged wind speed prediction. And the ANN model could beat both these two. Detailly information for errors of mean value model is given in Table 4. Compared the Table 4 with the results in Table 1 and Table 2, we have:

In 6-hour-averaged wind speed prediction, the RMSE of ANN model ($\sqrt{0.9384\ m^2/s^2} \approx 0.9687\ m/s$) is 47.45% less than that of the mean value model ($\sqrt{3.3977\ m^2/s^2} \approx 1.8433\ m/s$), the global MAE of ANN model (0.7279 m/s) is 48.40% less than that of the mean value model (1.4106 m/s).

In 24-hour-averaged wind speed prediction, RMSE of ANN model ($\sqrt{1.5482\ m^2/s^2} \approx 1.2443\ m/s$) is 14.78% less than that of the mean value model ($\sqrt{2.1317\ m^2/s^2} \approx 1.4600\ m/s$), the global MAE of ANN model (0.9457 m/s) is 15.65% less than that of the mean value model (1.1212 m/s).

**Table 4. statistical summary of MSE and MAE of mean value model for 6-hour-averaged and 24-hour-averaged wind speed prediction (unit: $m^2/s^2$)**

| | MSE | | | | MAE | | | |
| | 6-hour | | 24-hour | | 6-hour | | 24-hour | |
| | Sequence | Matrix | Sequence | Matrix | Sequence | Matrix | Sequence | Matrix |
|---|---|---|---|---|---|---|---|---|
| Max | 35.7526 | 11.1134 | 10.7192 | 7.1476 | 5.2803 | 2.6783 | 2.8390 | 2.1909 |
| Min | 0.4810 | 0.3911 | 0.3562 | 0.2663 | 0.5487 | 0.4961 | 0.4617 | 0.4065 |
| Median | 2.8552 | 3.9219 | 1.7063 | 2.4352 | 1.3373 | 1.5880 | 1.0402 | 1.2427 |
| Mean | 3.3977 | 3.3977 | 2.1317 | 2.1317 | 1.4106 | 1.4106 | 1.1212 | 1.1212 |
| Standard Deviance | 2.2948 | 1.7044 | 1.4081 | 1.0575 | 0.4324 | 0.4396 | 0.3517 | 0.3401 |

The relative error of mean value model is also larger than ANN model.

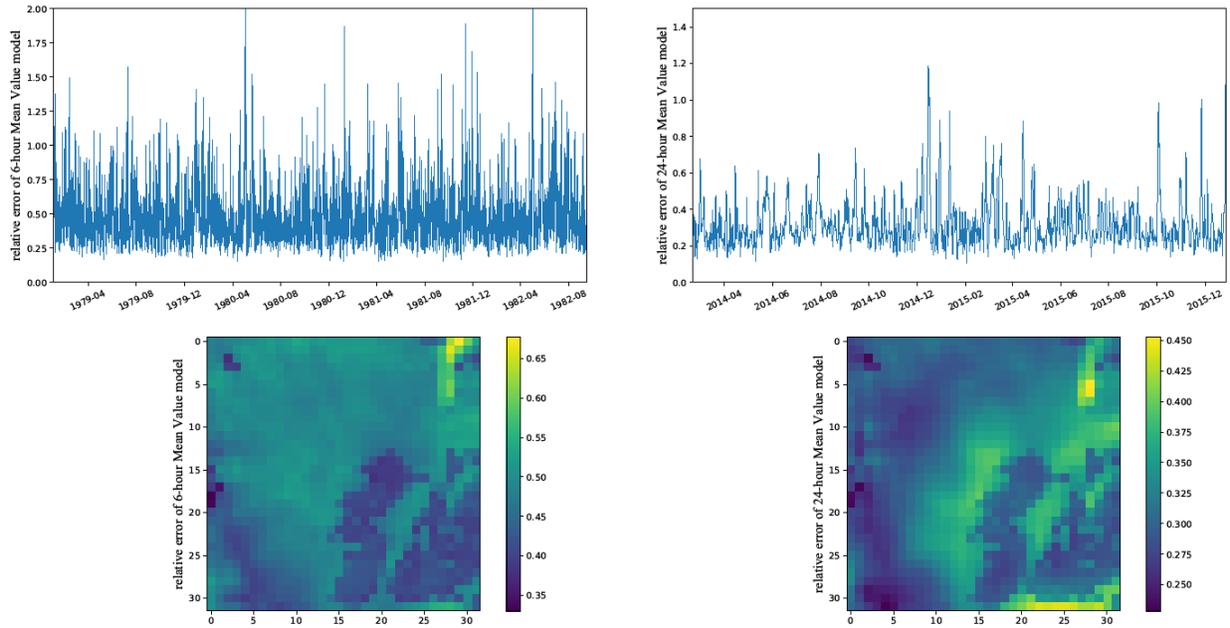

**Fig. 4.2.3 Comparison of mean relative error for mean value model in 6-hour-averaged (left) and 24-hour-averaged (right) wind speed prediction**

Table 5. Comparison of relative error between mean value model and ANN at different quantiles

|  | Relative Error | | | |
|---|---|---|---|---|
|  | 6-hour | | 24-hour | |
| Quantile | Mean Value | ANN | Mean Value | ANN |
| 0.025 | 0.0132 | 0.0063 | 0.0108 | 0.0087 |
| 0.25 | 0.1319 | 0.0654 | 0.1091 | 0.0889 |
| 0.5 | 0.2700 | 0.1435 | 0.2269 | 0.1887 |
| 0.75 | 0.4699 | 0.2706 | 0.3836 | 0.3363 |
| 0.975 | 2.4268 | 1.1508 | 1.2811 | 1.0766 |

### 4.3 ANN versus ARIMA

The computational cost of constructing an ARIMA for each spatial and temporal point in order to make a pointwise comparison with our ANN (like what we do in comparing persistence model with ANN) is very high since the size of our data set is very large (324336 hours and 1024 spatial points). Thus, to compare the performance of an ARIMA model to our ANN model, we select 5 different locations that represent locations where the errors of ANN are around quantiles 0, 0.25, 0.50, 0.75 and 1. We compared the performance of test errors between ANN and ARIMA at these 5 locations in the following manner.

At each location, we randomly select thousands of time points on test set of ANN as the ARIMA test points. Then before each test time point, we truncate a length of wind speed data at this location to construct the training time series and use BIC statistics on the training time series to select the best ARIMA models to give the forecast at that time point. We compare the ARIMA prediction error at each location and at each test time point with ANN error, and use statistical inference to test the difference of absolute errors and squared errors between these two kinds of models. The results show that the ANN error is less than ARIMA error from statistical perspective. For details, see **Appendix B.**

Unlike the persistence model and ANN model, ARIMA models are not so convenient for the wind speed simultaneous prediction in such a big area. On the one hand, since we have 1024 locations in our target area, if we want to use ARIMA, we need to train thousands of different models at the same time and then select the best ones to generate the prediction at each location respectively, which is much too expensive on time cost. On the other hand, the ANN models only need to be trained once. After the training process, we could forecast the wind speed 1-step ahead by loading the trained model directly, which can be very quick. However, if we use ARIMA, then every time we want to forecast, we need to train the models, which is very tedious.

## 5. Improvements and Confidence Interval for ANN Prediction

For a biased estimation $\hat{Y} = \hat{f}(X)$ for $Y$, we could generate an unbiased estimation $Z$ via:

$$Z = \hat{Y} + E[(Y - \hat{Y})] \tag{5.1}$$

where the term $E[(Y - \hat{Y})]$ represents the expectation of the bias. However, typically we don't know the joint distribution for $X$ and $Y$, so we do not have access to the expectation $E[(Y - \hat{Y})]$ for future unknown data (like test data). Fortunately, the errors on training set, test set, and validation set behaves very close (see **Appendix A.1**), which indicates that the performance of our ANN is stable enough such that we could approximately estimate the expectation of bias on test set by the sample mean on training set. This inspires us an improvement for our ANN prediction, i.e., generating an unbiased prediction $Z$ on test data via:

$$\hat{Z} = \hat{Y}_{test} + \hat{E}[(Y - \hat{Y}) \mid training\ set] \tag{5.2}$$

The first term on the RHS of Eq. (5.2) is the prediction achieved by ANN, the second term is computed by the training error of ANN.

Table 6. Comparison of MSE and MAE between biased ANN estimation $\hat{Y}$ and improved prediction $\hat{Z}$

|  |  |  | Training set | Test set | Validation set |
|---|---|---|---|---|---|
| MSE (unit: $m^2/s^2$) | 6-hour | $\hat{Y}$ | 0.8310 | 0.9384 | 0.9622 |
|  |  | $\hat{Z}$ | 0.8000 | 0.9086 | 0.9320 |
|  | 24-hour | $\hat{Y}$ | 1.4804 | 1.5482 | 1.5841 |
|  |  | $\hat{Z}$ | 1.4328 | 1.4977 | 1.5443 |
| MAE (unit: $m/s$) | 6-hour | $\hat{Y}$ | 0.6866 | 0.7279 | 0.9622 |
|  |  | $\hat{Z}$ | 0.6703 | 0.7132 | 0.7207 |
|  | 24-hour | $\hat{Y}$ | 0.9257 | 0.9457 | 0.9560 |
|  |  | $\hat{Z}$ | 0.9045 | 0.9248 | 0.9380 |

Assuming the errors are multivariate normal distributed, i.e.

$$Y - \hat{Y} \sim N(E[Y - \hat{Y}], \Sigma_0) \tag{5.3}$$

$$Y - Z \sim N(0, \Sigma_0) \tag{5.4}$$

and

$$(Y - Z)^T \Sigma_0^{-1} (Y - Z) \sim \chi_p^2 \tag{5.5}$$

Where $\Sigma_0$ is the covariance matrix of $Y - Z$.

Then we have the asymptotic approximation

$$\left(Y - \hat{Z}\right)^T \hat{\Sigma}^{-1} \left(Y - \hat{Z}\right) \xrightarrow{D} \chi_p^2 \tag{5.6}$$

Where $\hat{\Sigma}$ is the sample covariance matrix of training errors (shown in **Appendix A.4**), $\chi_p^2$ is the Chi-square distribution with degree of freedom $p$. We can use the asymptotic convergence relationship Eq. (5.6) to give the confidence interval. The confidence interval at level $\alpha$ for prediction $\hat{Z}$ is a high-dimensional ellipsoid region:

$$\left(Y - \hat{Z}\right)^T \hat{\Sigma}^{-1} \left(Y - \hat{Z}\right) \leq \chi_{p,1-\alpha}^2 \tag{5.7}$$

Where $\chi_{p,1-\alpha}^2$ is the $1 - \alpha$ quantile of the cumulative density function of $\chi_p^2$.

## 6. Summary and Discussion

In this paper, we proposed a composite Artificial Neural Network (ANN) which consists of a convolutional layer, a Long Short-Term Memory (LSTM) recurrent layer and a transposed convolutional layer. The ANN was trained to predict 1-step ahead 6-hour-averaged and 24-hour-averaged wind speed and the MSE and MAE were compared with persistence model, mean value model, and Integrated Autoregressive Moving Averaged (ARIMA) model. The ANN model has three distinct advantages which 1) could predict the wind speed in a large area simultaneously, 2) perform a relative high accuracy, and 3) only need to be trained once and then we could put it into use. The persistence model and mean value model also have the advantages 1) and 3) but the accuracy is obviously lower than ANN. The accuracy of ARIMA models is close to ANN but do not have the advantages 1) and 3).

Predictions of full wind fields are necessary for operators who must constantly balance the supply to meet the demand on the grid and particularly useful for future power planning such as the optimization of electricity power supply systems. And the ANN we proposed here could give the decision maker more accurate prediction and more reliable suggestions.

# Appendix

## A. Error Analysis on ANN models

### A.1 Expectation and Standard Deviance of ANN errors

As we mentioned before, because of the regularization term in the loss function, the ANN estimation is not an unbiased estimation. The error of our ANN model $\hat{f}(x)$ is:

$$\hat{\varepsilon} = Y - \hat{Y} = \left(f(x) - \hat{f}(x)\right) + \varepsilon \tag{A.1.1}$$

So

$$E(\hat{\varepsilon}) = E[f(x) - \hat{f}(x)] \tag{A.1.2}$$

$$SD(\hat{\varepsilon}) = \sqrt{Var(\hat{\varepsilon})} = \sqrt{Var\left(f(x) - \hat{f}(x)\right) + E(\varepsilon^2)} \tag{A.1.3}$$

We use sample mean $\hat{E}(\hat{\varepsilon})$ and sample standard deviance $\widehat{SD}(\hat{\varepsilon})$ to estimate $E(\hat{\varepsilon})$ and $SD(\hat{\varepsilon})$. Because of the regularization, the sample estimations are very stable on training set, test set, and validation set, as shown from Fig. A.1 to Fig. A.4.

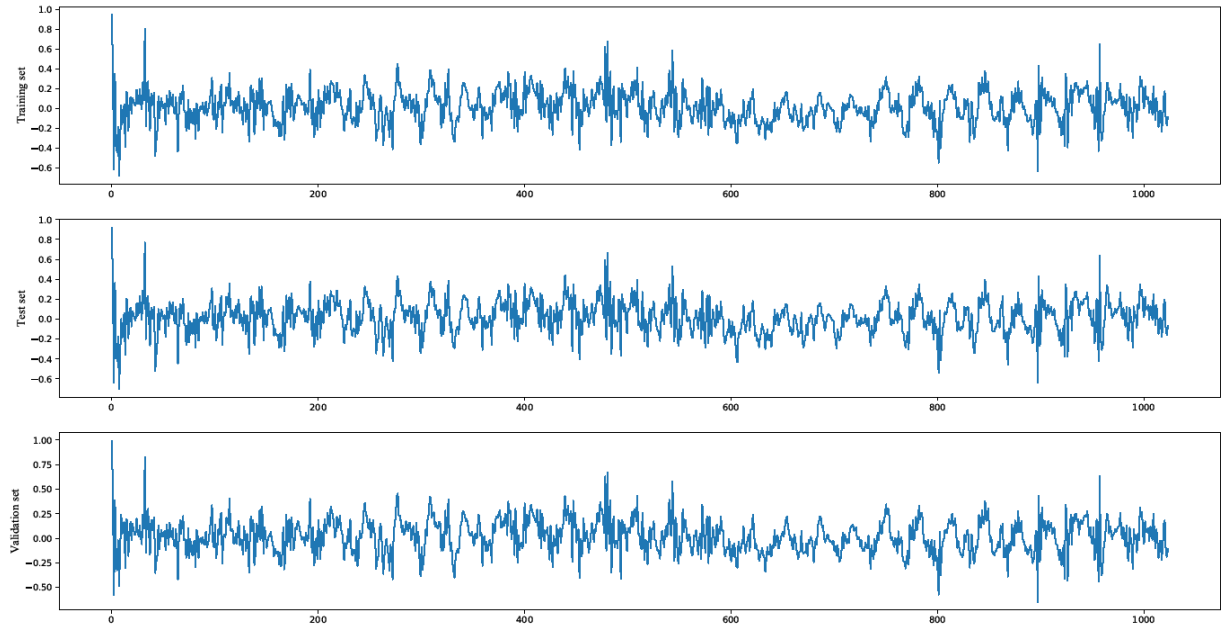

**Fig. A.1 Sample averaged error $\hat{E}(\hat{\varepsilon})$ at 1024 locations for 6-hour-averaged wind speed prediction on each data set**

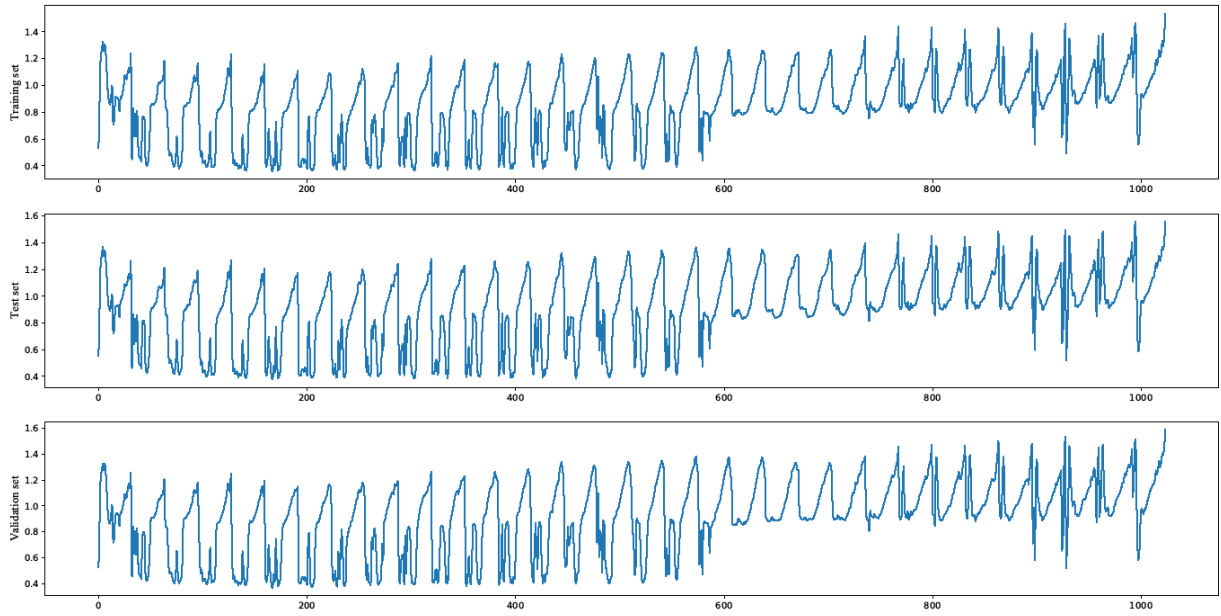

**Fig. A.2 sample standard deviance of error $\widehat{E}(\hat{\varepsilon}^2)$ at 1024 locations for 6-hour-averaged wind speed prediction on each data set**

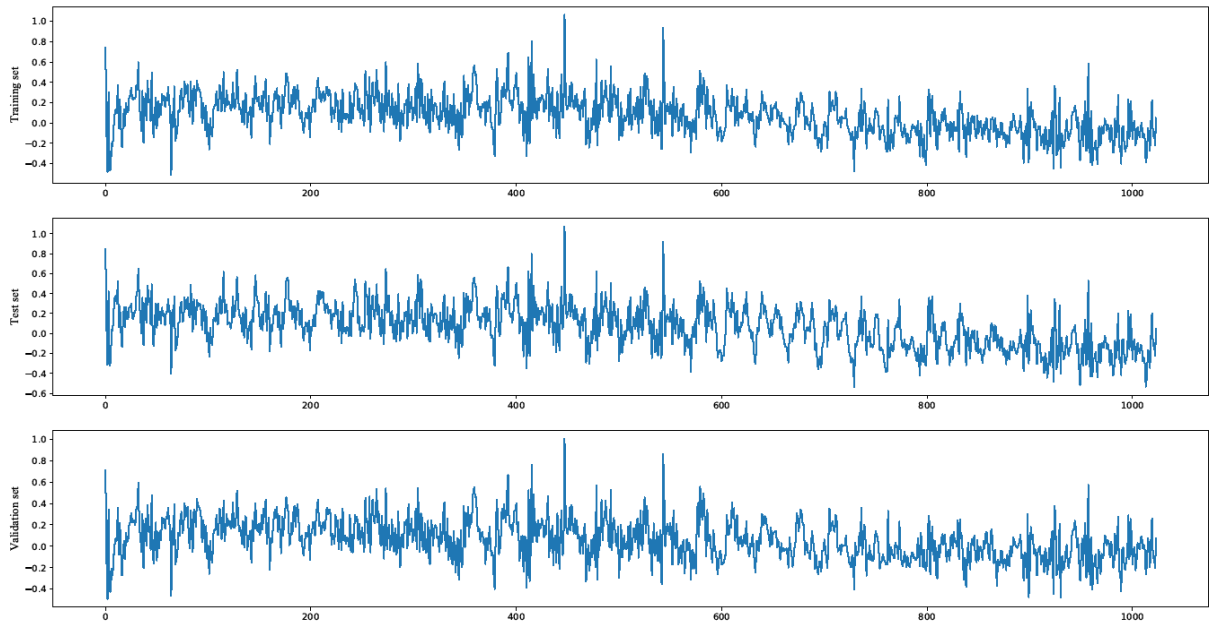

**Fig. A.3 Sample averaged error $\widehat{E}(\hat{\varepsilon})$ at 1024 locations for 24-hour-averaged wind speed prediction on each data set**

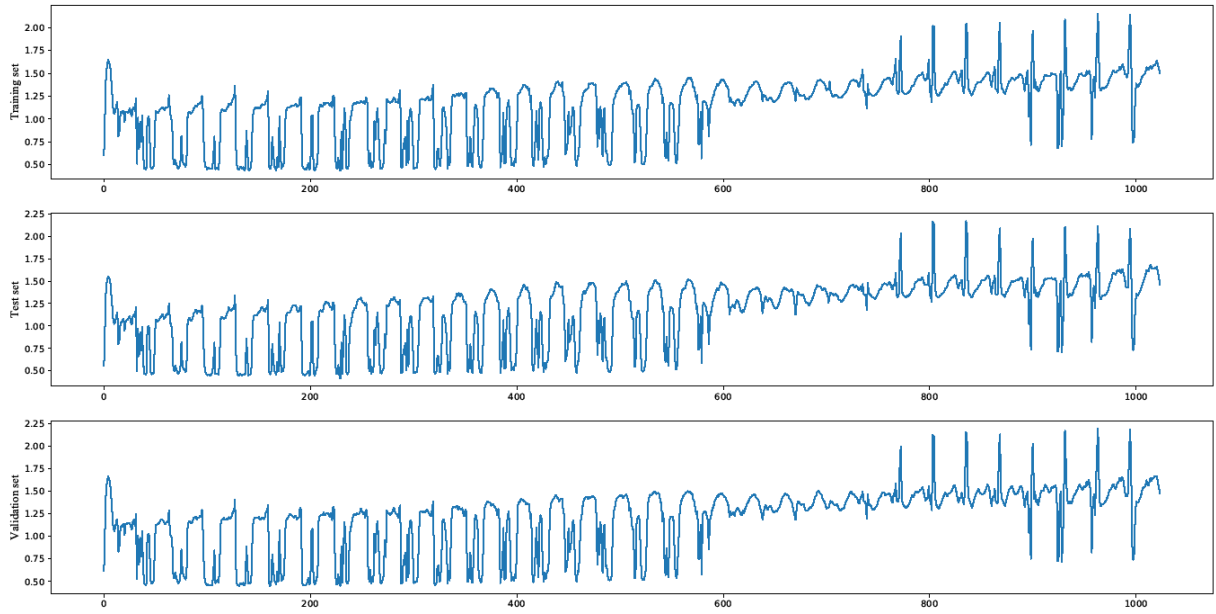

**Fig. A.4 sample standard deviance of error $\widehat{E}(\hat{\varepsilon}^2)$ at 1024 locations for 24-hour-averaged wind speed prediction on each data set**

## A.2 Seasonality and Topographic Dependence of ANN MSE

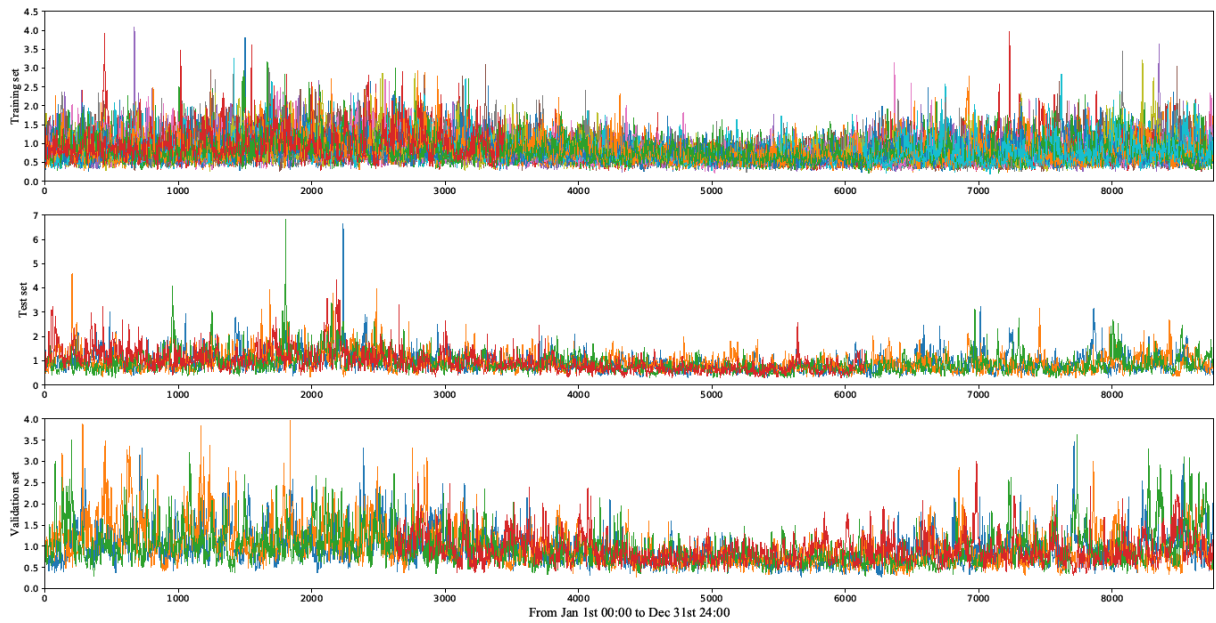

**Fig. A.5 MSE sequence of ANN model for 6-hour-averaged wind speed prediction among each data set**

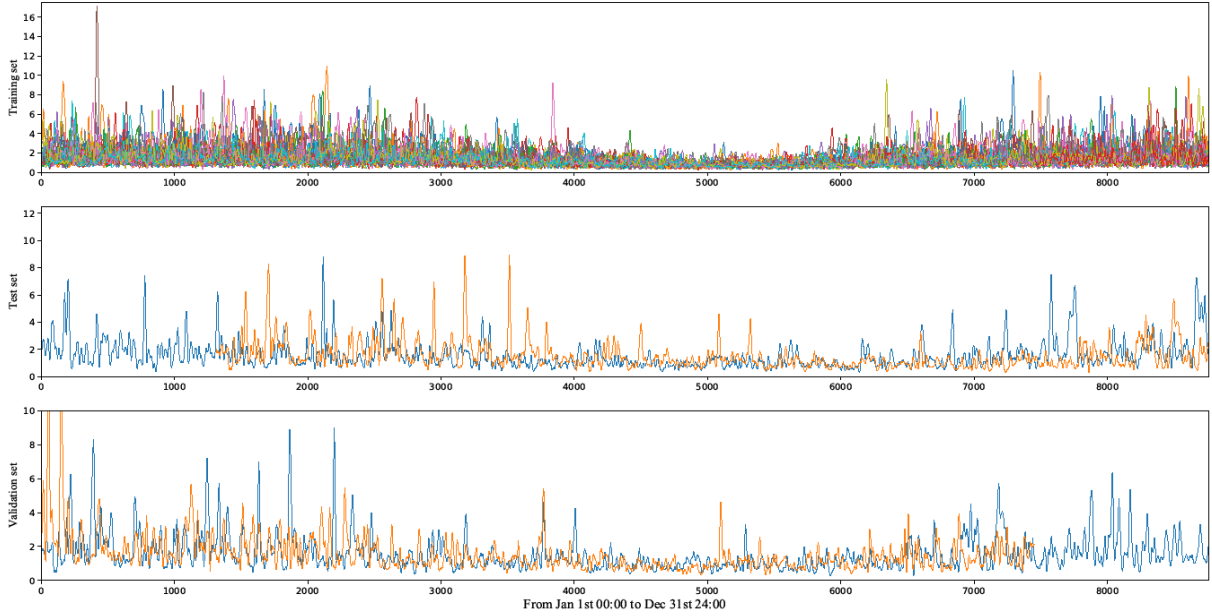

**Fig. A.6 MSE sequence of ANN model for 24-hour-averaged wind speed prediction among each data set**

The equation (A.1.1) tells us that the seasonality and the spatial correlation of our prediction error $\hat{\varepsilon}$ comes from the first term $f(x) - \hat{f}(x)$. This term is dependent on the wind speed $x$.

Some clues can be found. Fig. A.5 and Fig. A.6 show that the spatial mean squared error goes to the lowest level during JJA season, which is in accordance with the wind speed magnitude (Fig. 2.1.2). From Fig. 4.1.1 to Fig. 4.1.4, we could see that the MSE (MAE) of ANN has a strong dependence on topography while the similar relationship between wind speed magnitude and topography can also be observed in Fig. 2.1.2.

## A.3 ACF of ANN error

Besides the expectation and variance of errors, we also checked the autocorrelation of ANN errors. In short term, we computed the autocorrelation of errors at each spatial point in 4 steps (at lag 6, 12, 18, 24 hours for 6-hour model, at lag 24, 48, 72, 96 hours for 24-hour model).

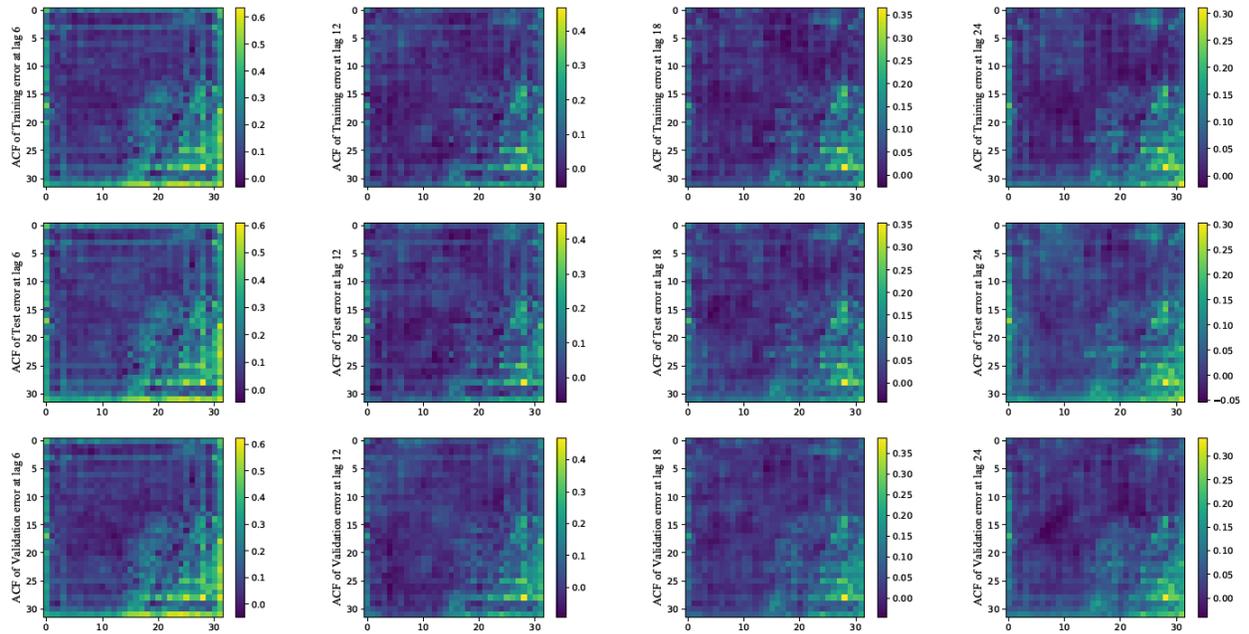

**Fig. A.7** autocorrelation of ANN errors on each data set for 6-hour-averaged wind speed prediction at lag 6h, 12h, 18h, 24h

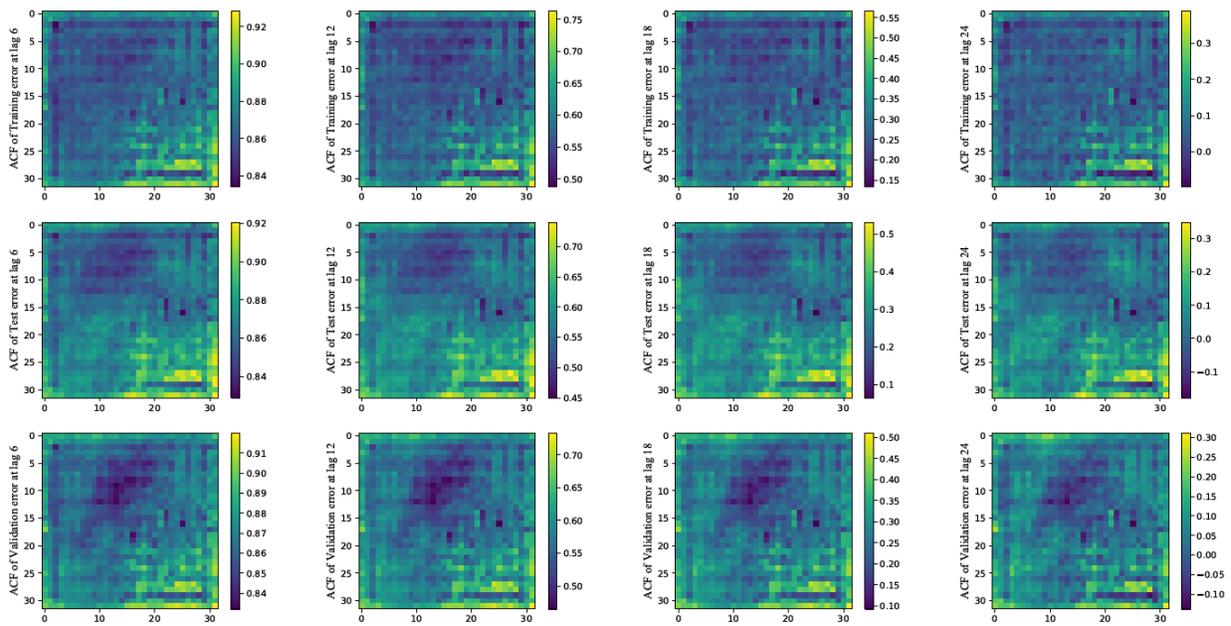

**Fig. A.8** autocorrelation of ANN errors on each data set for 24-hour-averaged wind speed prediction at lag 24h, 48h, 72h, 96h

To check the time dependence in long-term, we computed the autocorrelation of spatial mean error from 1024 locations. The time lag is from 0 to 3 years and to 1 year respectively for 6-hour-averaged and 24-hour-averaged wind speed prediction.

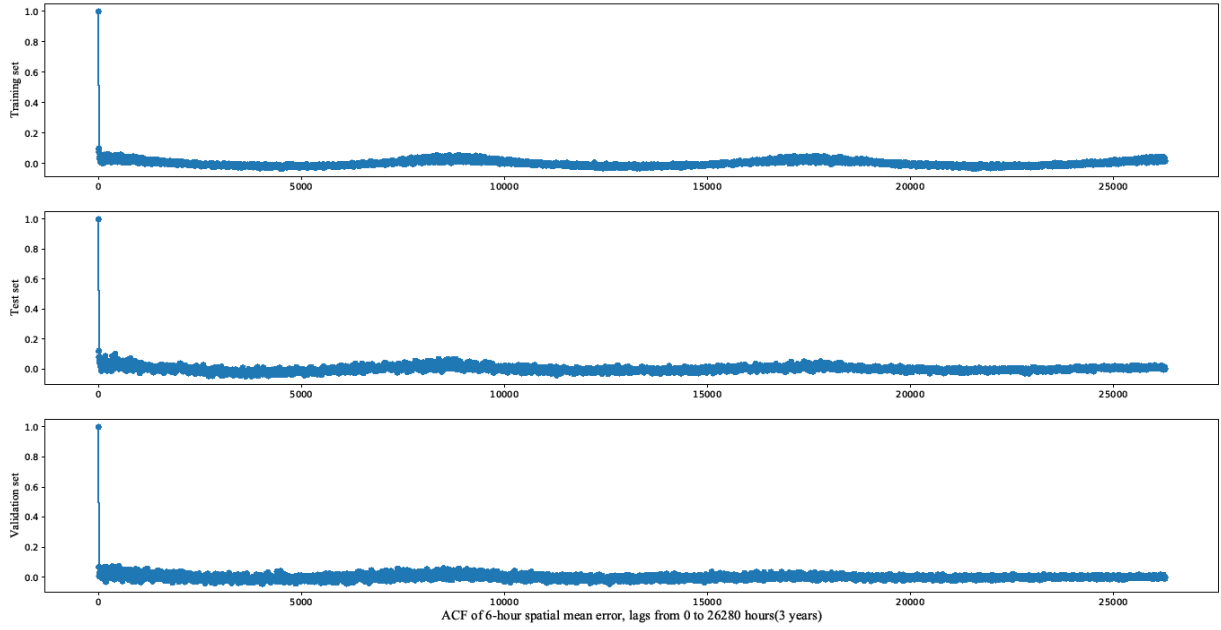

**Fig. A.9 autocorrelation of spatial mean error on each data set for 6-hour-averaged wind speed prediction**

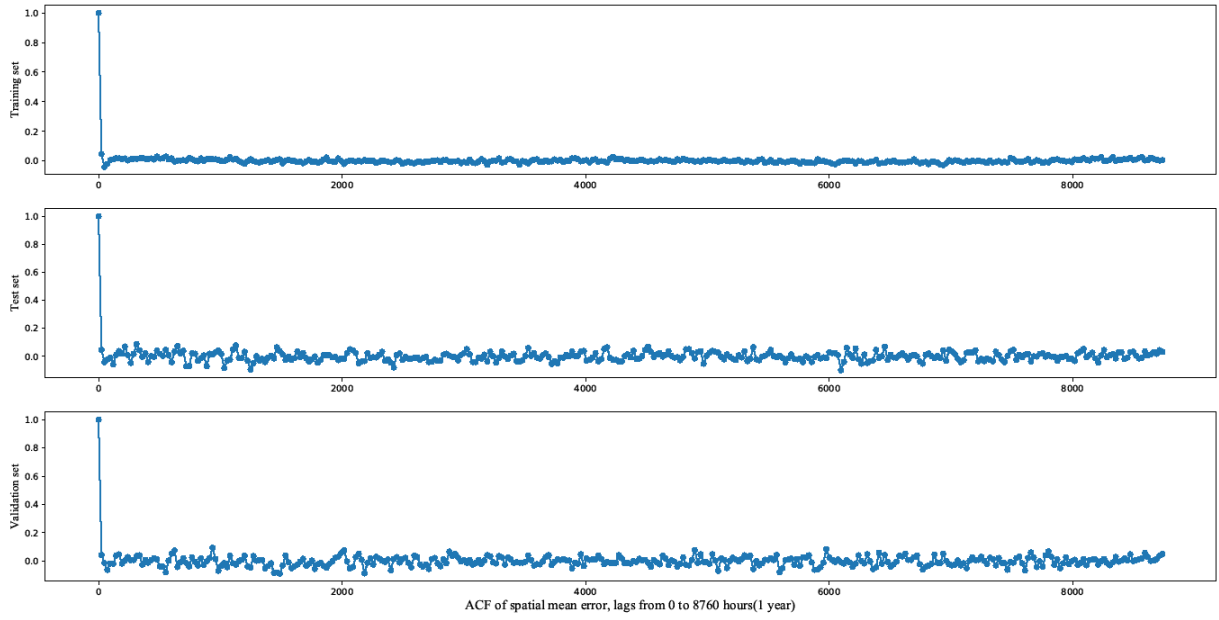

**Fig. A.10 autocorrelation of spatial mean error on each data set for 6-hour-averaged wind speed prediction**

## A.4 Spatial correlation of ANN error

Wind speed at two close points has strong correlation (Fig. 2.1.5), so does the error of ANN. We ordered the 1024 spatial points by row (the most northwest point as the $0^{th}$, and the most southeast point as the $1023^{rd}$), then computed the correlations between the ANN errors at these points as shown below.

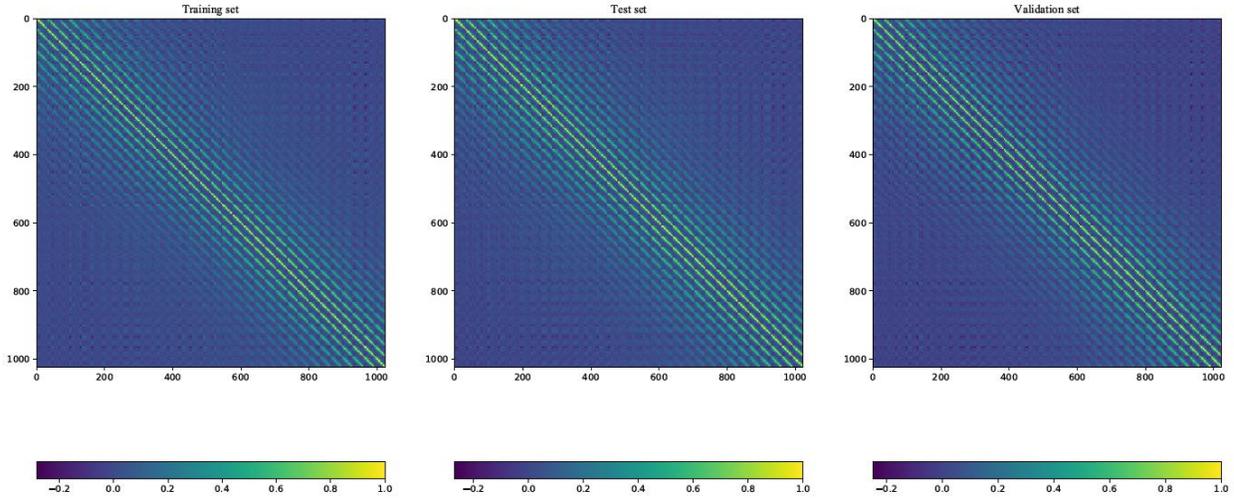

Fig. A.11 Correlation matrix of ANN errors at 1024 spatial points for 6-hour-averaged wind speed prediction

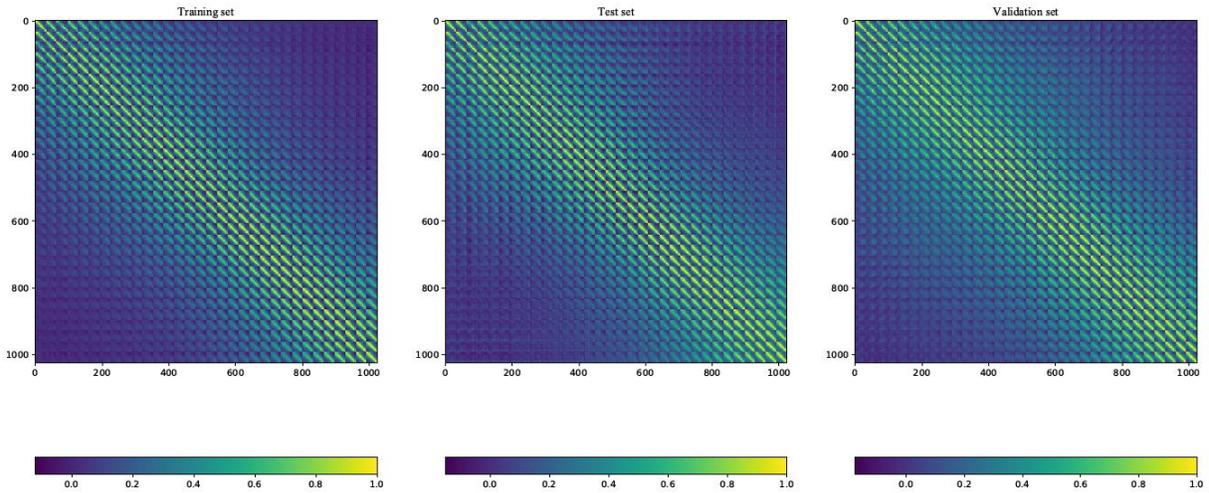

Fig. A.12 Correlation matrix of ANN errors at 1024 spatial points for 24-hour-averaged wind speed prediction

We plotted the histograms of these matrixes to check the sparsity:

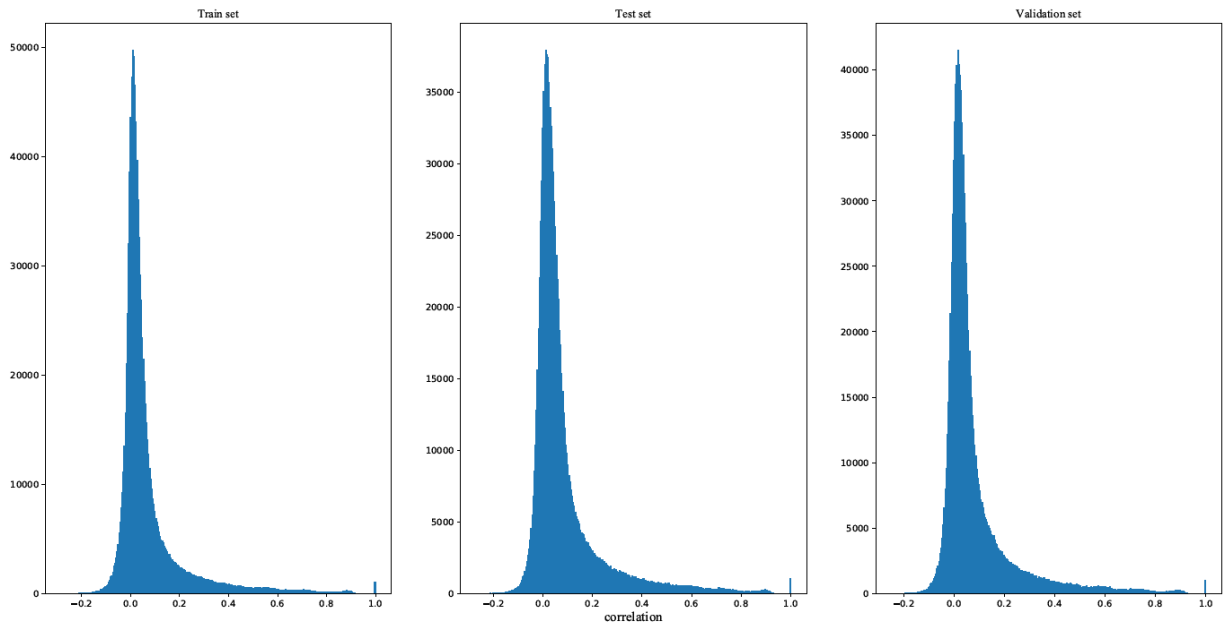

**Fig. A.13 Histogram of correlation matrix of 6-hour ANN errors**

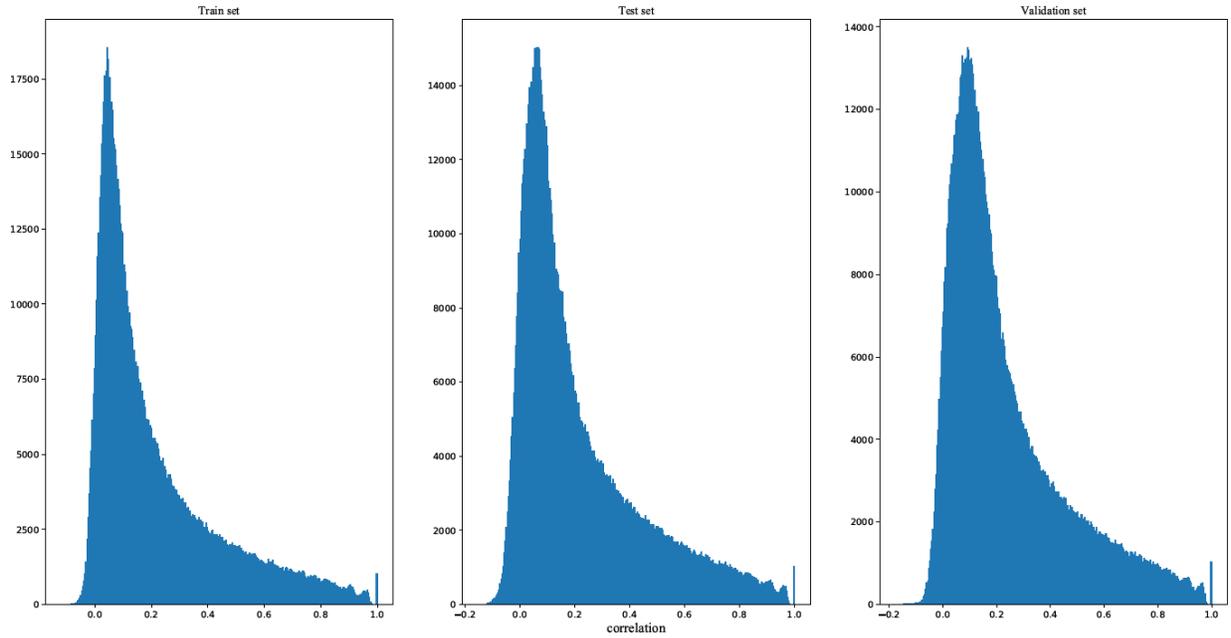

**Fig. A.14 Histogram of correlation matrix of 24-hour ANN errors**

## B. ARIMA errors versus ANN errors

The length of test set for 6-hour-averaged wind speed prediction is 3.7 years. We chose the last 36 months and randomly selected 40 time points as the test points in each month (thus 36×40=1440 points in total) to compute the ARIMA test errors. We truncated 6000 hours before each test time point to construct 1000-long training time series and found the best ARIMA (p, d, q) model using the training BIC statistics under the difference time d=0 and d=1 respectively. We generate the prediction at these test points using the best ARIMA models we found and compare with the ANN, and use statistical inference to test the difference of absolute errors and squared errors between these two kinds of models.

We use the sample mean $\hat{\mu}$ and sample standard deviance $\hat{\sigma}$ to test if the ANN error is smaller than the ARIMA error, i.e., whether we can reject the null hypothesis:

$$\mu = Error(ANN) - Error(ARIMA) = 0$$

Assuming normal distribution, the 95% confidence interval for $\mu$ is:

$$\left(\hat{\mu} - 1.96 \cdot \hat{\sigma}/\sqrt{n}, \hat{\mu} + 1.96 \cdot \hat{\sigma}/\sqrt{n}\right)$$

Similar works have been done in 24-hour-averaged wind speed prediction. Choose the last 22 months in the ANN test set, in each month randomly select 60 time points (thus 22x60=1320 points in total) to construct ARIMA models at these points.

**Difference time d=0:**

Table B.1 statistical inference for the difference of absolute errors $|error_{ANN}| - |error_{best\ ARIMA}|$ and squared errors $|error_{ANN}|^2 - |error_{best\ ARIMA}|^2$.

| | | Quantile | $\hat{\mu} - 1.96 \cdot \hat{\sigma}/\sqrt{n}$ | $\hat{\mu} + 1.96 \cdot \hat{\sigma}/\sqrt{n}$ | $sign(\mu)$ |
|---|---|---|---|---|---|
| Difference of Absolute Errors | 6-hour | 0 | -0.0344105 | -0.0055895 | − |
| | | .25 | -0.03078799 | 0.02478799 | 0 |
| | | .5 | 0.04275143 | 0.10524857 | + |
| | | .75 | -0.05881673 | 0.00481673 | 0 |
| | | 1 | -0.11220276 | -0.03379724 | − |
| | 24-hour | 0 | -0.02299733 | 0.00699733 | 0 |
| | | .25 | -0.14373283 | -0.09226717 | − |
| | | .5 | -0.23919859 | -0.16680141 | − |
| | | .75 | -0.22322754 | -0.15277246 | − |
| | | 1 | -0.58689456 | -0.45310544 | − |
| Difference of Squared Errors | 6-hour | 0 | -0.03557982 | -0.00242018 | − |
| | | .25 | -0.07163616 | 0.02763616 | 0 |
| | | .5 | 0.08821586 | 0.22178414 | + |
| | | .75 | -0.13094649 | 0.04094649 | 0 |
| | | 1 | -0.34843468 | -0.11156532 | − |
| | 24-hour | 0 | -0.0364539 | -0.0035461 | − |
| | | .25 | -0.33322615 | -0.20677385 | − |
| | | .5 | -0.74294572 | -0.49705428 | − |
| | | .75 | -0.73147729 | -0.47652271 | − |
| | | 1 | -3.27623446 | -2.47576554 | − |

**Difference time d=1:**

Table B.2 statistical inference for the difference of absolute errors $\mu = |error_{ANN}| - |error_{best\ ARIMA}|$ and the difference of squared errors $\mu = |error_{ANN}|^2 - |error_{best\ ARIMA}|^2$

|  |  | Quantile | $\hat{\mu} - 1.96 \cdot \hat{\sigma}/\sqrt{n}$ | $\hat{\mu} + 1.96 \cdot \hat{\sigma}/\sqrt{n}$ | $sign(\mu)$ |
|---|---|---|---|---|---|
| Difference of Absolute Errors | 6-hour | 0 | -0.04280654 | -0.01119346 | − |
|  |  | .25 | -0.0500486 | 0.0100486 | 0 |
|  |  | .5 | -0.10076701 | -0.02923299 | − |
|  |  | .75 | -0.07108674 | 0.00508674 | 0 |
|  |  | 1 | -0.13974725 | -0.04825275 | − |
|  | 24-hour | 0 | -0.01499733 | 0.01499733 | 0 |
|  |  | .25 | -0.11547623 | -0.05052377 | − |
|  |  | .5 | -0.24041042 | -0.16358958 | − |
|  |  | .75 | -0.22664726 | -0.14335274 | − |
|  |  | 1 | -0.55523533 | -0.41076467 | − |
| Difference of Squared Errors | 6-hour | 0 | -0.04623416 | -0.00976584 | − |
|  |  | .25 | -0.0633788 | 0.0453788 | 0 |
|  |  | .5 | 0.04042389 | 0.19557611 | + |
|  |  | .75 | -0.17052472 | 0.03652472 | 0 |
|  |  | 1 | -0.45315094 | -0.16884906 | − |
|  | 24-hour | 0 | -0.02950785 | 0.00350785 | 0 |
|  |  | .25 | -0.34378004 | -0.17621996 | − |
|  |  | .5 | -0.72823255 | -0.47176745 | − |
|  |  | .75 | -0.76079563 | -0.44520437 | − |
|  |  | 1 | -3.21153571 | -2.36046429 | − |

Where $sign(\mu)$ :

$'+': Error(ANN) > Error(ARIMA)$,
$'-': Error(ANN) < Error(ARIMA)$,

$'0': could\ not\ reject\ the\ null\ hypothesis\ Error(ANN) = Error(ARIMA)\ at\ level\ 0.05$

## 6-hour models:

## Difference time d=0:

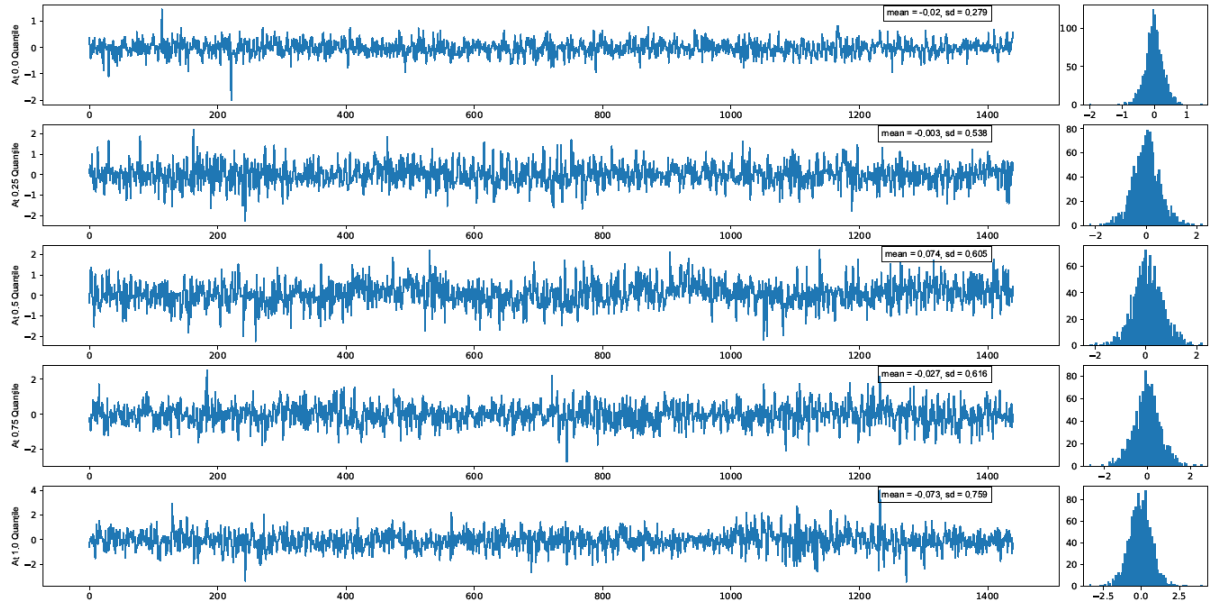

**Figure B.1** Comparison between absolute error at 1440 time points at five spatial locations: $|error_{ANN}| - |error_{best\ ARIMA}|$

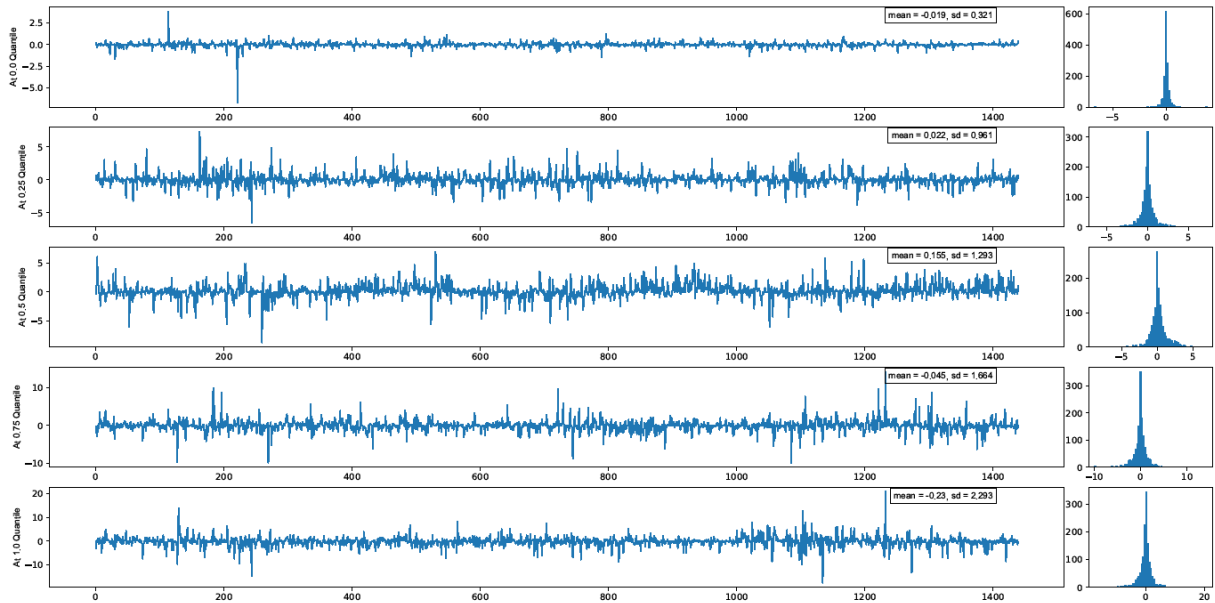

**Figure B.2** Comparison between squared error at 1440 time points at five spatial locations: $|error_{ANN}|^2 - |error_{best\ ARIMA}|^2$

**Difference time d=1:**

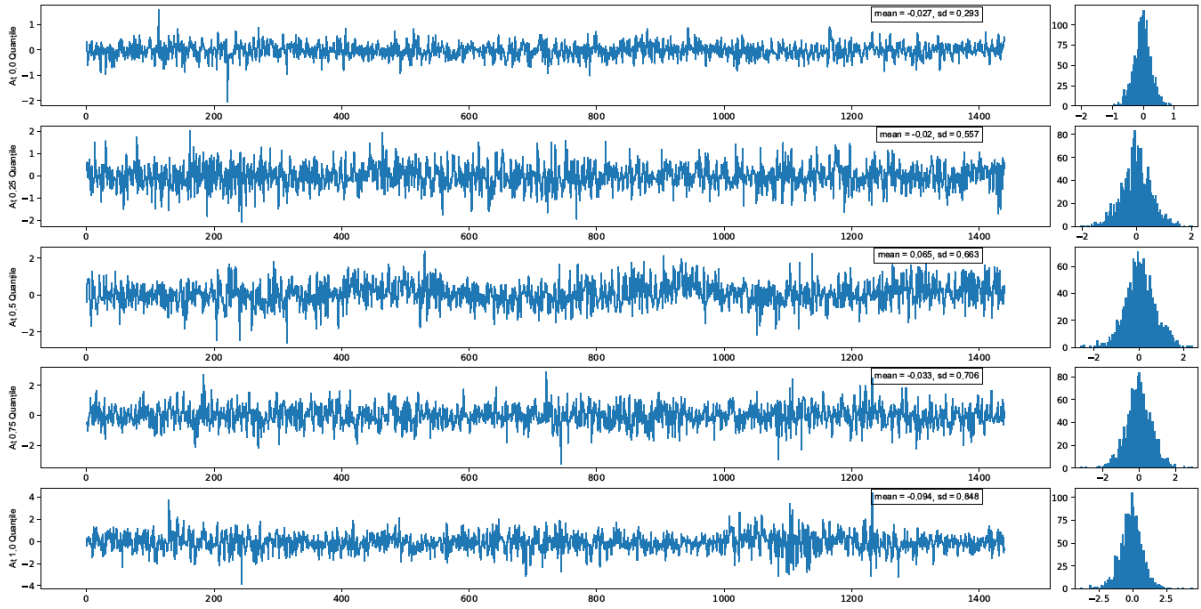

**Figure B.3** Comparison between absolute error at 1440 time points at five spatial locations: $|error_{ANN}| - |error_{best\ ARIMA}|$

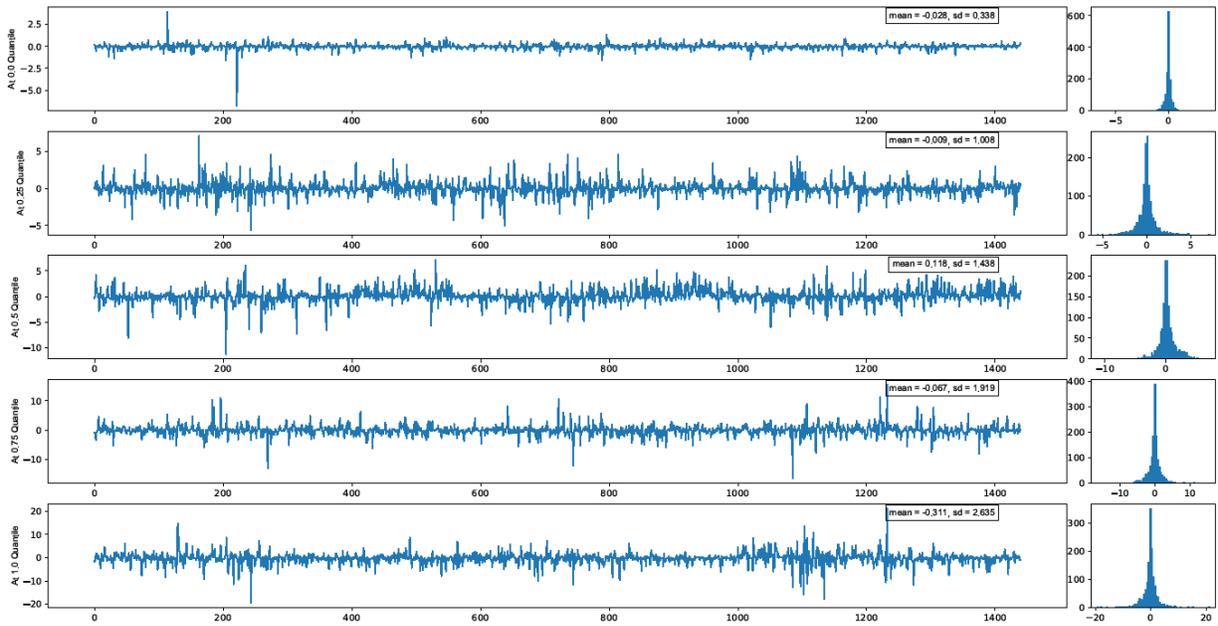

**Figure B.4** Comparison between squared error at 1440 time points at five spatial locations: $|error_{ANN}|^2 - |error_{best\ ARIMA}|^2$

## 24-hour models:

## Difference time d=0:

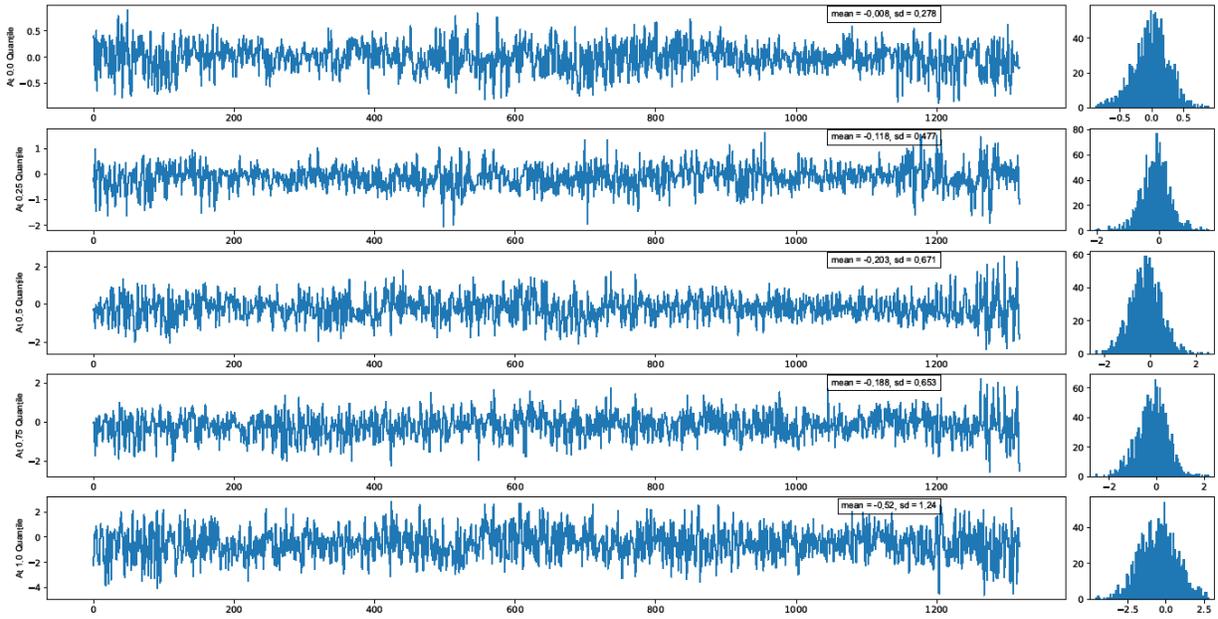

**Figure B.5** Comparison between absolute error at 1320 time points at five spatial locations: $|error_{ANN}| - |error_{best\,ARIMA}|$

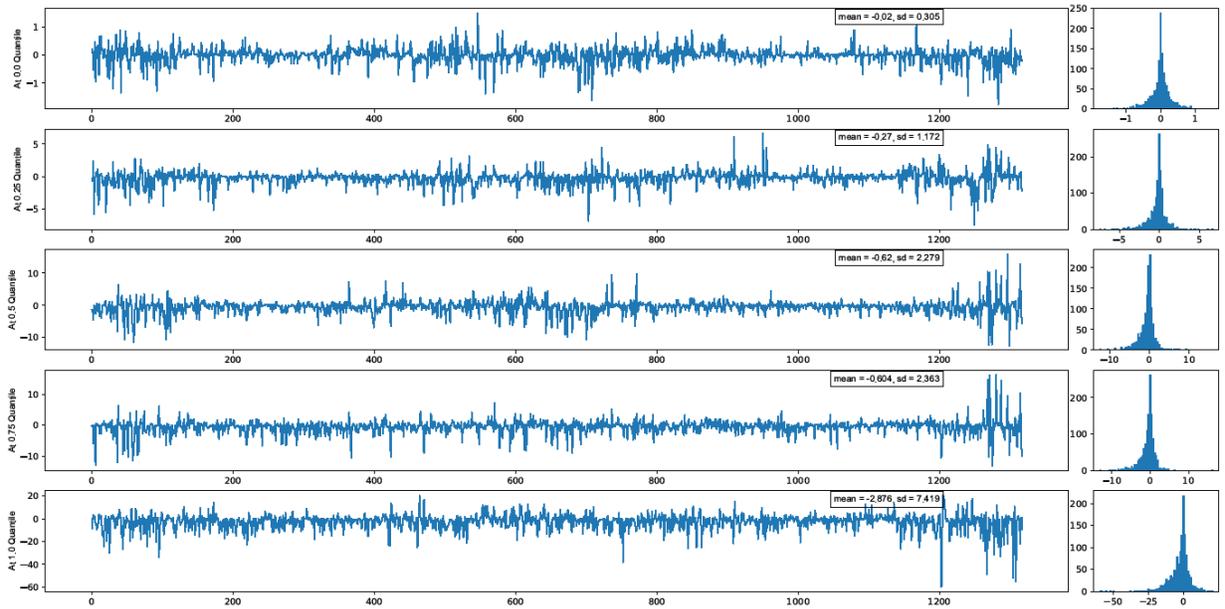

**Figure B.6** Comparison between squared error at 1320 time points at five spatial locations: $|error_{ANN}|^2 - |error_{best\,ARIMA}|^2$

**Difference time d=1:**

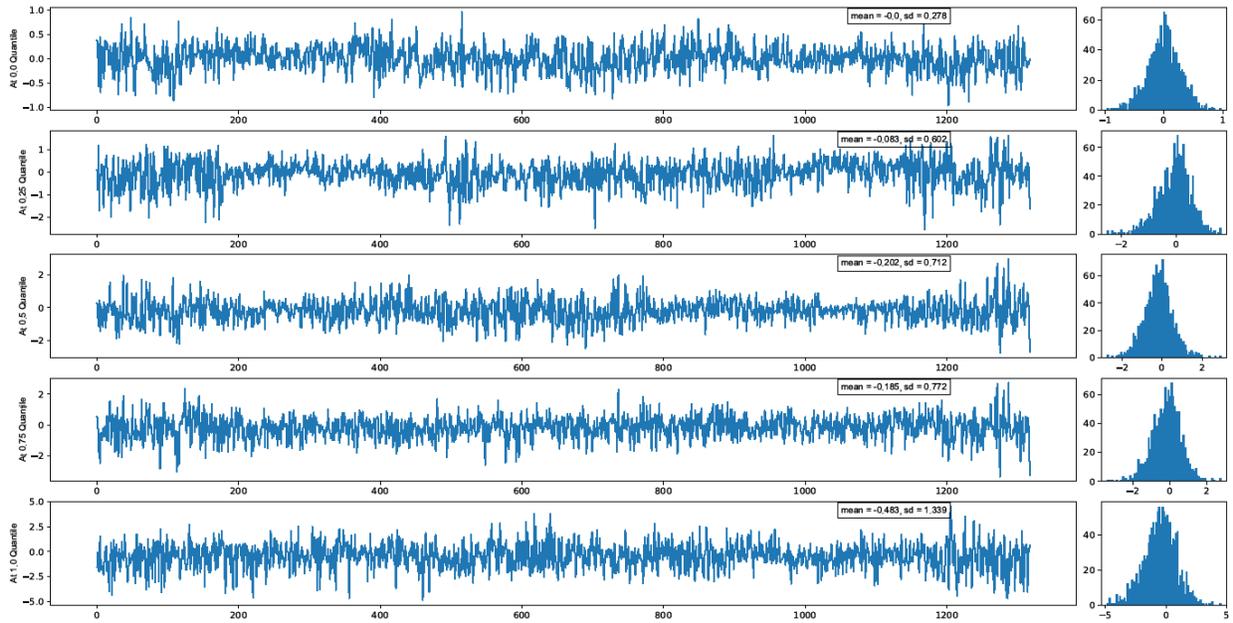

Figure B.7 Comparison between absolute error at 1320 time points at five spatial locations: $|error_{ANN}| - |error_{best\ ARIMA}|$

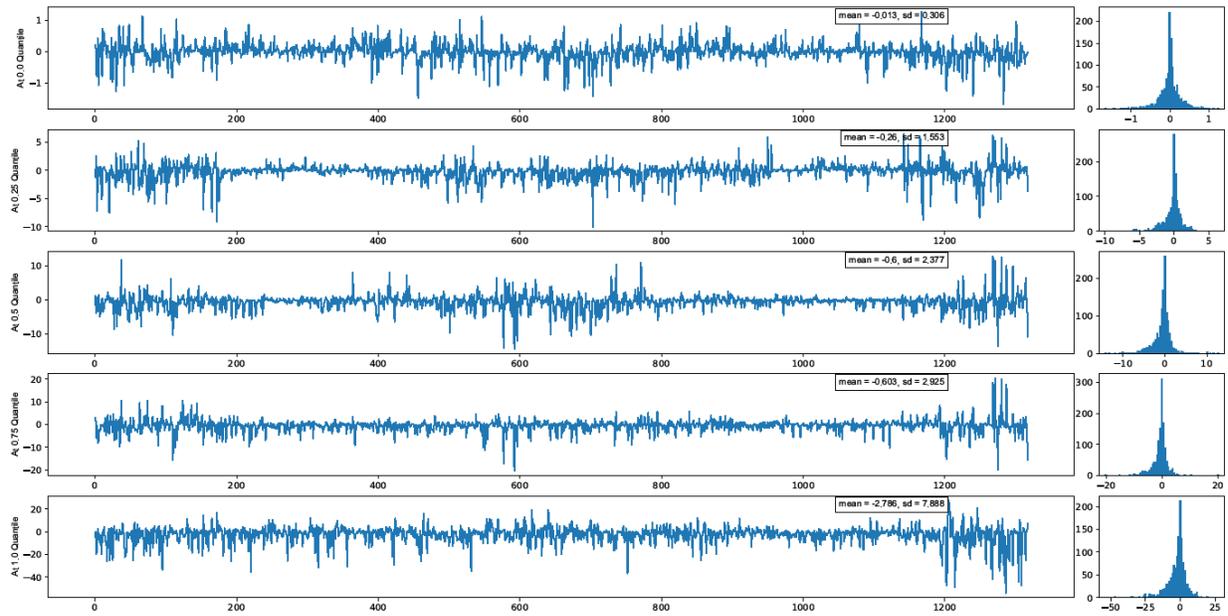

Figure B.8 Comparison between squared error at 1320 time points at five spatial locations: $|error_{ANN}|^2 - |error_{best\ ARIMA}|^2$

**C. Choice for Hyper-Coefficients**

The choice of $n$ (the length of single input series) for 6-hour-averaged and 24-hour-averaged wind speed prediction is actually a result of trade-off between model complexity and data sufficiency: due to the constraints of the RAM and computation power of our computing machine, as the $n$ increases, the model itself could learn longer-time information, but the number of training series has to decrease, which leads to data insufficiency. The $n$ chosen here is by experience and experimental results. We could not prove it is the best choice since there are some other hyper-coefficients (e.g. the regularization number $\lambda$) to tune, but under the constraints of computation power and data amount, the $n$ we use here is at least a suboptimal choice.

The other hyper-coefficients, such as the regularization number in ANN and the length of training time series in ARIMA model fitting, are also chosen by experimental results. In view of the limited space, not describe detailly here.